\documentclass[]{fairmeta}

\usepackage{hyperref}
\usepackage{url}
\usepackage{enumitem}
\usepackage{booktabs}
\usepackage{graphicx}
\usepackage{tcolorbox}
\usepackage{multirow} 
\usepackage[table]{xcolor}
\usepackage{wrapfig}
\usepackage{amssymb}

\title{Adaptive Personalization Under Preference Epistemic Uncertainty}

\author[1,2,*]{Priyanka Dey}
\author[1]{Rajarshi Das}
\author[1]{Vikas Bahirwani}
\author[1]{Qi Zeng}
\author[1]{Julia Kruk}
\author[1]{Qi Qi}

\affiliation[1]{Meta Superintelligence Labs}
\affiliation[2]{University of Southern California}

\contribution[*]{Work done at Meta}

\abstract{
Human preferences change over time: an established preference may remain valid over time, become uncertain as context shifts, or be explicitly revised. 
Effective personalization requires conversational AI to adapt its response strategy to evolving preferences rather than relying on passive memory retrieval. 
We formalize this capability as \textit{adaptive personalization under epistemic uncertainty}. 
A human preference study further confirms that people favor state-contingent response strategies as preference uncertainty changes. 
To evaluate whether language models mirror this adaptation, we create a multilingual benchmark spanning 250 personas, 5 languages, and 25K conversational turns that decouples strategy selection from response execution. 
Across eight closed- and open-weight models, we uncover a \textit{recognition--generation gap}: models reliably identify appropriate strategies when selecting among candidate options, but collapse during open-ended generation, suffering a 35-46\% drop in strategy calibration. 
Crucially, this failure is systematic: models default to exploit-first behavior, over-relying on existing preferences regardless of preference epistemic uncertainty.
We show this failure is addressable: while prompting interventions fail, targeted training that supervises models' internal strategy representations toward state-conditional human preferences narrows the user-acceptability generalization gap by a third on out-of-distribution personas and conversational styles. 
}

\correspondence{Priyanka Dey at \email{deyp@usc.edu}}

\metadata[Code]{\url{https://github.com/priyanka-dey/adaptive-personalization}}

\begin{document}

\maketitle


\section{Introduction}

\begin{figure*}[!h]
\centering
\includegraphics[width=\textwidth]{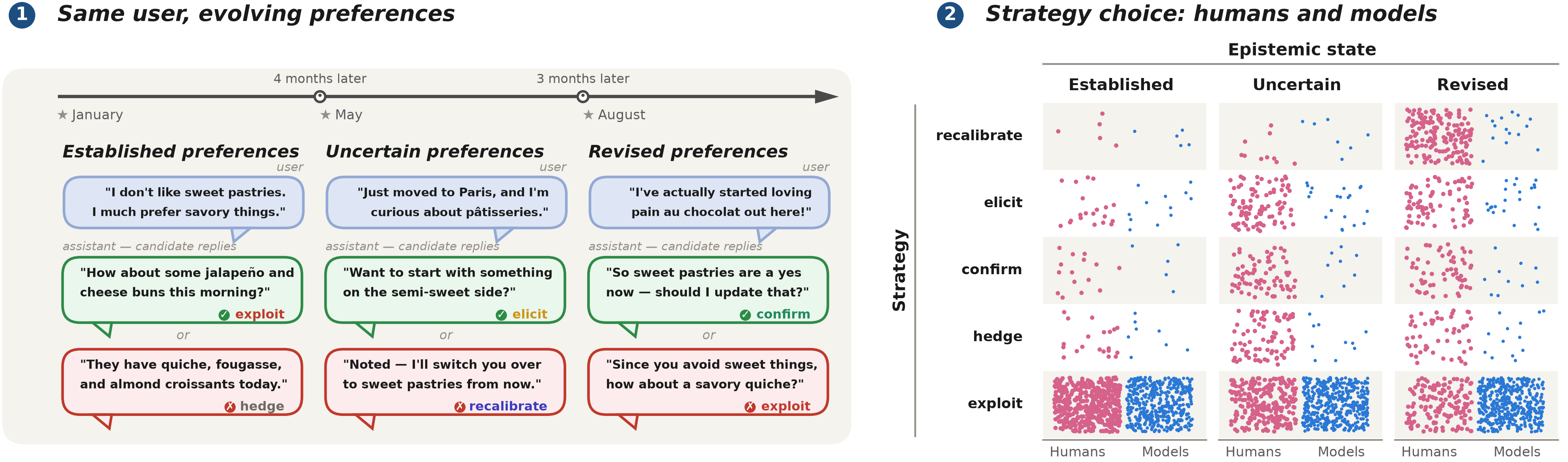}
\vspace{-0.5em}
\caption{\textbf{Humans prefer varying response strategies based on
preference epistemic uncertainty; models exploit regardless.}
\textbf{(Left)} One user at three time points, illustrating the three
preference epistemic states we evaluate: 1) \emph{established}, 2)
\emph{uncertain}, and 3) \emph{revised}, each paired with a
well-matched and a mismatched assistant reply. \textbf{(Right)}
Across the three epistemic states (preference uncertainties), humans prefer responses with varying strategies, whereas current models predominantly \emph{exploit} in all states.}
\label{fig:teaser}
\vspace{-0.5em}
\end{figure*}

Human preferences evolve dynamically across interactions: an established preference may remain valid over time, become uncertain as context shifts, or be explicitly revised by new evidence. For conversational AI, adapting to these shifts requires far more than passive memory retrieval. It demands deciding whether a remembered preference should still be acted upon, questioned, hedged around, or updated. Consider the user in Figure~\ref{fig:teaser}, who states a preference for savory food, later mentions exploring local bakeries, and eventually says she has come to enjoy something sweet, without ever restating her preferences. An assistant that stubbornly mentions her preference for savory foods across all three turns has retrieved the user’s history correctly, but failed to adjust the conversational strategy to match the evolving reliability of that preference.

We formalize this capability as \emph{adaptive personalization under epistemic uncertainty}. Evidence regarding user preferences evolves across three distinct epistemic uncertainty states: \textbf{established} (explicitly stated and uncontradicted), \textbf{uncertain} (ambiguous or partial context), and \textbf{revised} (superseded by new evidence). Each state demands a distinct conversational behavior: acting directly on settled preferences, seeking clarification under ambiguity \cite[dialogue grounding;][]{clark1996using,clark1991grounding,traum1994computational}, or course-correcting when preferences change \cite[belief revision;][]{alchourron1985logic,gardenfors1988knowledge,hansson1999belief}. A controlled human study confirms this state-contingent behavior: humans systematically prefer distinct model response strategies across epistemic states (\S\ref{sec:human_study}). 

Despite human demand for dynamic strategy adaptation, existing personalization benchmarks predominantly evaluate model retrieval of preferences over long-horizon interactions \citep{jiang2025know, zhao2025prefeval, jiang2025personamem, harry2026beyond, guo2026realpref}, but fail to test whether conversational strategy adapts to preference reliability over time. To measure this distinct capability, we create 
a benchmark spanning 250 personas, five languages, and 25,006 conversational turns, with oracle epistemic-state labels and an evaluation framework that decouples strategy selection from response execution (\S\ref{sec:task}). 

Evaluating eight closed- and open-weight models demonstrates that models’ failures localize to generation rather than recognition (\S\ref{sec:results}). Across varying epistemic states, models select responses with appropriate strategies at a high accuracy when presented with candidate options, but fail to generate responses – a gap widening to over 35\% under epistemic states: uncertain and revised. Moreover, models collapse to an exploit-first default generation policy that over-relies on historical preferences regardless of preference epistemic uncertainty.


To address this default behavioral policy in models, we first test inference-time prompting interventions (\S\ref{sec:interventions}); however, directly providing preference context fails to close the recognition-generation gap. Internal probing demonstrates that although models represent epistemic states in their hidden activations, this passive knowledge fails to govern downstream strategy selection for generation (\S\ref{sec:training}). Motivated by this diagnostic, we train models using a reward function that aligns predicted models’ response strategies with human preference choices, conditioned on preference epistemic uncertainty (\S\ref{sec:training}). This targeted training narrows the user-acceptability generalization gap by a
third to out-of-distribution personas and conversational styles beyond standard fine tuning.

Together, these results identify adaptive personalization under epistemic uncertainty as a distinct, learnable capability: one that contemporary language models do not reliably deploy by default, but one that targeted training can build.

\section{Task Formulation and Benchmark}
\label{sec:task}
\subsection{Task Formulation}

Dialogue theories characterize conversation as a process of establishing and maintaining common ground, in which interlocutors continuously distinguish between mutually established information and information that remains uncertain or ungrounded \citep{clark1996using,clark1991grounding}. Formal theories of belief revision describe how such beliefs should be updated when new evidence conflicts with existing knowledge
\citep{alchourron1985logic,gardenfors1988knowledge,hansson1999belief}. We ground our task in these two areas and adapt them to a personalization setting, where an assistant's evidence about a user's preferences can be established, uncertain, or revised, and its conversational strategy should track that evidence. 

An assistant's evidence for a preference (epistemic uncertainty) is \textit{established} when explicitly expressed and uncontradicted, \textit{uncertain} when evidence is insufficient or ambiguous, or \textit{revised} when newer evidence supersedes it. 
Given these varying epistemic states, communication theory, belief revision theory and linguistic studies \citep{clark1996using,clark1991grounding, lakoff1973hedges, alchourron1985logic,gardenfors1988knowledge} distinguishes several strategies an assistant can take in conversation:
\begin{enumerate}
    \item \textit{exploit}: acting directly on a stated preference
    \item \textit{elicit}: asking the user for missing information
    \item \textit{confirm}: checking whether a preference still holds before acting on it
    \item \textit{hedge}: offering options without fully committing to either
    \item \textit{recalibrate}: updating behavior once new evidence contradicts a prior preference
\end{enumerate}
Together, these five strategies form an operational set of conversational actions (Table~\ref{tab:strategies}; full definitions in \S~\ref{app:benchmark_generation}). 
Formally, at each turn the assistant observes a dialogue context $x$ with epistemic state $s \in \{\textsc{established}, \textsc{uncertain}, \textsc{revised}\}$ over a preference and must select an appropriate conversational strategy $a \in \{\textsc{exploit},\textsc{elicit}, \textsc{confirm}, \textsc{hedge}, \textsc{recalibrate}\}$.

\subsection{Preferred Strategies Depend on Epistemic State}
\label{sec:human_study}

We conducted a multilingual preference study across five languages - English, Spanish, Portuguese, Hindi, and Indonesian, to measure which assistant response strategies humans prefer across different epistemic states. Annotators were presented with dialogues exhibiting varying epistemic states and chose their preferred response between LLM outputs implementing different conversational strategies. The study included 500 dialogue instances evaluated by 3 annotators per item ($\sim$1,500 total annotations).
Table~\ref{tab:human_pref} reports this resulting preference distribution.

\begin{wraptable}{r}{0.42\columnwidth}
\vspace{-10pt}
\centering
\scriptsize
\caption{\textbf{Preferred strategy varies systematically with preference epistemic state.} \checkmark marks the set of acceptable strategies used for benchmark construction, evaluation, and training.}
\label{tab:human_pref}
\begin{tabular}{lccc}
\toprule
\textbf{Strategy} & \textbf{Established} & \textbf{Uncertain} & \textbf{Revised} \\
\midrule
Exploit      & 90.4\%\checkmark & 8.5\%              & 6.2\%             \\
Elicit       & 2.4\%            & 27.0\%\checkmark   & 20.0\%\checkmark   \\
Confirm      & 2.0\%            & 28.4\%\checkmark   & 24.6\%\checkmark   \\
Hedge        & 4.0\%            & 26.6\%\checkmark   & 23.0\%\checkmark   \\
Recalibrate  & 1.2\%            & 9.5\%             & 26.2\%\checkmark   \\
\bottomrule
\end{tabular}
\vspace{-10pt}
\end{wraptable}

Human preferences are not evenly distributed across the five strategies: in each epistemic state, a small set of strategies account for the vast majority of preferences ($>80\%$) (Table~\ref{tab:human_pref}), suggesting a natural separation between acceptable and unacceptable strategies. To validate this separation, we exhaustively evaluated all 15 possible two-group splits of the five strategies in each state using a $\chi^2$ goodness-of-fit test against the observed human choices (\S~\ref{app:human_validation}). In every state, one split fit the observed preferences best.
We therefore use these best-fitting splits to define our human-grounded sets of \emph{acceptable} and \emph{unacceptable} strategies: \textit{exploit} for \textit{established}; \textit{elicit, confirm, hedge} for \textit{uncertain}; and \textit{elicit, confirm, hedge, recalibrate} for \textit{revised}.

\subsection{Benchmark Construction}
\label{sec:benchmark_construct}

Evaluating strategy selection under preference uncertainty requires two properties that existing real-world dialogue corpora lack: controlled, multi-session preference trajectories with known ground-truth transitions, and counterfactual candidate responses spanning multiple strategies at each turn. To address this gap, we introduce a multilingual benchmark spanning five languages (English, Spanish, Portuguese, Hindi, and Indonesian; Figure~\ref{fig:benchmark}), constructed through a three-role multi-LLM pipeline.

Starting from 250 diverse personas seeded from PersonaHub \cite{tencent2024personahub} using farthest-point sampling, each persona is augmented with preference attributes drawn from 20 life domains \cite{personalens} and 13 interaction styles \cite{li2026horizonbench}. A \emph{user simulator} plans multi-session conversation arcs (10-14 sessions) with oracle-labeled preference trajectories, systematically injecting planned drift events that transition a preference between the three epistemic states: \textit{established}, \textit{uncertain}, and \textit{revised}. An \emph{AI assistant} then generates candidate responses across strategies at each turn, while an \emph{LLM jury} filters output for conversation coherence and label fidelity.

Our final benchmark comprises 25,006 turns across 2,939 multi-session conversations. Each turn includes four candidate LLM responses: two acceptable strategies (per Table~\ref{tab:human_pref}) and two negative foils categorized into seven realistic failure modes identified during preliminary annotation (over-personalization, epistemic overreach, staleness, need mismatch, over-probing, style mismatch, and failure to probe; see \S~\ref{app:benchmark_generation} for full statistics and construction details).

We validate our benchmark through a human evaluation following established benchmark protocols \cite{zollo2025personalllm, jang2026scizoom, dey2025can}. Human annotators evaluated 500 dialogue instances across all five languages (3 annotators per language) on five dimensions: conversation coherence, epistemic-state clarity, strategy-label correctness, foil-type accuracy, and response acceptability. Agreement between human annotators and the LLM jury was substantial ($\kappa = 0.78$; \S~\ref{app:human_validation}), while human multiple-choice accuracy (selecting a response with acceptable strategy use from candidate responses) reached 92\%. These results indicate that the benchmark's epistemic states and strategy labels are both recognizable to humans and reflected in natural dialogue.

\subsection{Evaluation Protocol}
\label{sec:eval}

At each turn, given persona $P$, conversation history $C$, and current user turn $u$, the benchmark provides four candidate responses $(r_1^+, r_2^+, r_3^-, r_4^-)$, where $r_1^+$ and $r_2^+$ implement acceptable strategies for the current epistemic state, while $r_3^-$ and $r_4^-$ implement foil strategies (\S~\ref{sec:task}). To benchmark \emph{adaptive personalization under epistemic uncertainty}, we evaluate models across two settings: \textit{multiple-choice selection} (MCQ), where the model selects one candidate from $\{r_1^+, r_2^+, r_3^-, r_4^-\}$, and \textit{free-response generation} (FRG), where the model generates an unguided response given $(P, C, u)$. Both settings are evaluated on \textbf{strategy calibration (SC)}: whether the chosen or generated response utilizes an acceptable strategy\footnote{When responses combine strategies, we score the dominant strategy: the strategy appearing most strongly in response. This avoids counting strategy-stacking responses as calibrated across states. Alternative scoring rules in (Appendix~\ref{app:scoring_rules}).} for the epistemic state (MCQ: matching $r_1^+$ or $r_2^+$ with a 50\% chance baseline; FRG: classified as acceptable via an LLM jury; \S~\ref{sec:benchmark_construct}). Comparing SC across settings explicitly isolates strategy recognition from strategy generation.

FRG outputs are additionally evaluated along two dimensions by the user simulator (\S~\ref{sec:benchmark_construct}): a state-conditioned \textbf{rubric (R)} scoring five binary criteria (out of 5): Information, Confidence, Intent, Scope, and Style—and \textbf{user acceptability (UA)}, a binary accept/reject judgment. Together, SC, R, and UA decouple strategy selection, response execution, and overall user satisfaction.

\section{LLMs Fail to Adapt Conversational Strategy Under Epistemic Uncertainty}
\label{sec:results}

\S\ref{sec:human_study} showed that people prefer varying conversational strategies based on epistemic state. 
We now use our multi-lingual benchmark to measure whether models adapt strategy during conversation generation. 
We evaluate eight closed- and open-weight models (GPT-5.4, Opus 4.7, Grok 4.1, Gemini 3 Pro, Gemma 3 12B, Llama-3.1-8B, OLMo-3-7B, and Qwen3-8B) across the three epistemic states and across five languages.

\subsection{Strategy Calibration Collapses When Preference Evidence Becomes Uncertain or Revised}
\label{sec:collapse}

\begin{wrapfigure}{r}{0.5\textwidth}
\centering
\includegraphics[width=\linewidth]{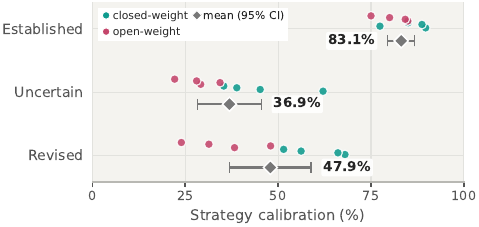}
\caption{\textbf{Strategy calibration across epistemic states.} All
models perform best in \textit{Established} and struggle with \textit{Uncertain} and \textit{Revised} states.}
\label{fig:main}
\end{wrapfigure}

We first measure whether models adapt their open-ended generation strategies based on epistemic state. Figure~2 shows free-response (FRG) strategy calibration (SC) by epistemic state, i.e., whether a response's dominant strategy is acceptable for that state. Calibration is high when preferences are established ($\sim83.1\%$ across models) but declines significantly once conversations require managing uncertainty ($\sim36.9\%$; best model: GPT-5.4 at $62\%$) or adapting to a revision ($\sim47.9\%$; best: GPT-5.4 at $68\%$). Both declines are significant after accounting for the nested structure of the data (mixed-effects logistic regression with random intercepts for persona and conversation; Uncertain: $\beta = -2.16$, Revised: $\beta = -1.67$, both $p < .001$; full specification and per-language estimates in \S~\ref{app:statistics}). This trend holds across models and languages, and is deeper in Hindi and Indonesian. closed-weight models retain higher calibration than open-weight models, and this margin widens with epistemic difficulty: 4.2 points when preferences are established, 16.9 under uncertainty, and 25.3 under revision.

\subsection{Models Recognize Appropriate Strategies More Often Than They Generate Them}
\label{sec:ladder}

This decline in SC could reflect either of two distinct failures: a \textit{generation} failure, in which models correctly \textit{recognize} the appropriate strategy but fail to produce it, or a more fundamental \textit{recognition} failure, in which models cannot correctly identify the appropriate strategy at all.
We distinguish these failure modes using our two evaluation methods: multiple-choice questions (MCQ), where a model must select the appropriate response (text following acceptable strategy) among candidates, and free-response generation (FRG), which requires the model to generate text following the acceptable strategy. 

\begin{figure}[t]
\centering
\includegraphics[width=\linewidth]{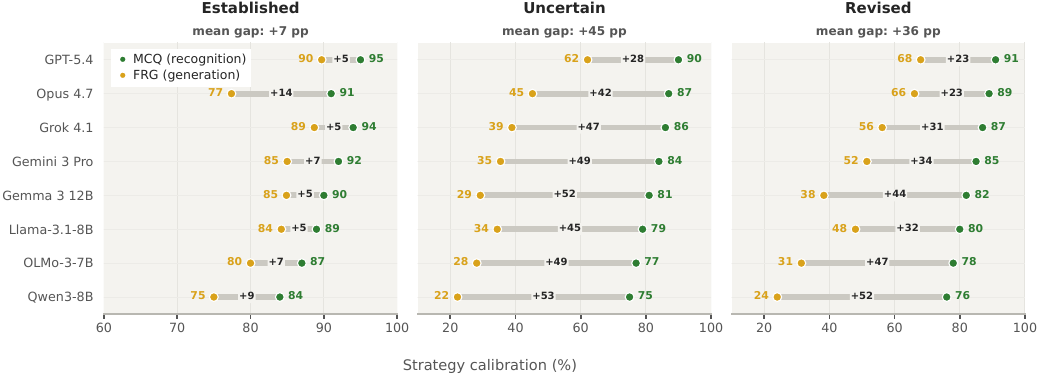}
\caption{\textbf{Recognition exceeds generation.} Models identify acceptable strategies (MCQ, green) more often than they generate them (FRG, yellow) across all three epistemic states. The horizontal distance between paired points is the recognition-generation gap.}
\label{fig:gap}
\end{figure}

Figure~\ref{fig:gap} compares the two evaluation results. MCQ strategy calibration remains high across all three states (mean 85.4\%, range: 75-95\%; 50\% chance baseline), declining only modestly under uncertainty and revision relative to established preferences (90.3\% $\to$ 82.4\% $\to$ 83.5\%). 
FRG strategy calibration, by contrast, collapses far more sharply across the three states (83.1\% $\to$ 36.9\% $\to$ 47.9\%; \S\ref{sec:collapse}). 
The resulting recognition--generation gap is present for every model in every state, and is widest under \textsc{Uncertain} ($+45$ pp, $p<.001$), followed by \textsc{Revised} ($+36$ pp, $p<.001$) and \textsc{Established} ($+7$ pp, $p < .001$).
Models can therefore identify the appropriate strategy substantially more reliably than they produce it: the failure characterized in \S\ref{sec:collapse} is not explained by an inability to recognize appropriate behavior.
A further check, explicitly supplying the correct strategy and asking models only to execute it, shows this gap is driven specifically by selecting a strategy, not executing one: models generate responses with the supplied strategy on 95\% of turns, and execution quality under this condition is state-invariant, whereas unaided calibration still swings by over 40 points across states (\S~\ref{app:oracle_strategy})

\subsection{Generation Failures Reflect Systematic Over-Exploitation}
\label{sec:over_exploit}

\begin{figure*}[t]
\centering
\includegraphics[width=\textwidth]{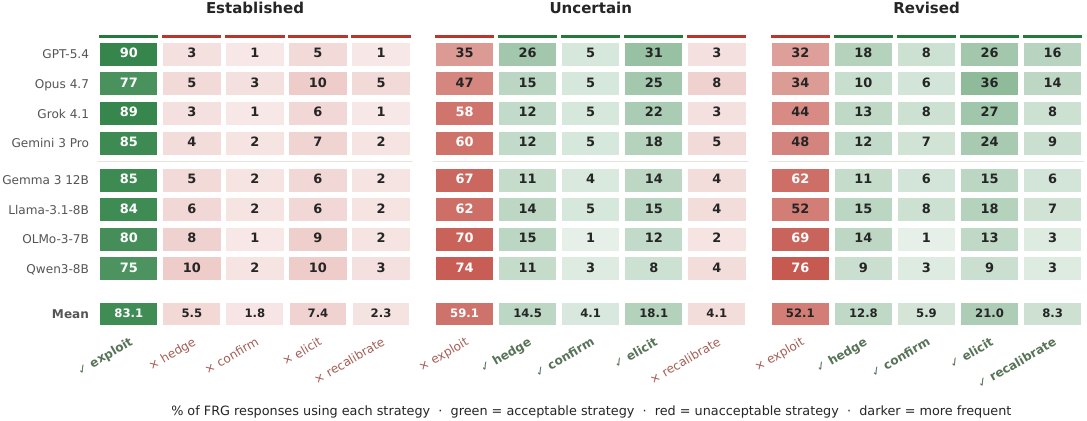}
\caption{\textbf{Models default to exploit across epistemic states.}
Each cell shows the percentage of free-response generations for which a strategy is dominant, pooled across five languages. Green indicates acceptable strategies and red indicates unacceptable strategies. \textit{Exploit} is the most frequent dominant strategy in 23 of 24 model$\times$state cells, including 59\% and 52\% of responses under \textsc{Uncertain} and \textsc{Revised}, respectively, where it is unacceptable. Multi-label strategy presence is reported in Figure~\ref{fig:app_presence}.}
\label{fig:heatmap}
\end{figure*}

In this section, we characterize models' FRG outputs by their dominant
strategy (Figure~\ref{fig:heatmap}). Averaged across models, \textit{exploit}
is the dominant strategy in 83\%, 59\%, and 52\% of responses under
\textsc{Established}, \textsc{Uncertain}, and \textsc{Revised}. Models do shift
toward acceptable strategies as evidence weakens (e.g., dominant
\textit{elicit} rises from 7\% to 18\% and 21\%), but far too little:
\textit{exploit} remains the most frequent dominant strategy in 23 of 24
model$\times$state cells, the sole exception being Opus-4.7 under
\textsc{Revised} (\textit{elicit} 36\% vs.\ 34\%). Exploitation is also
near-universal as a component of responses, present in 90\%, 91\%, and 86\% of
responses across states (Figure~\ref{fig:app_presence}), often
alongside a question or hedge. Models are therefore not selecting strategies
at random, but defaulting to the same policy, \textit{exploit}, regardless of
whether it remains appropriate.

\begin{wraptable}{r}{0.4\textwidth}
\centering
\small
\setlength{\tabcolsep}{3pt}
\caption{\textbf{FRG performance}, Avg. across epistemic states and languages w/ 95\% CI. SC = strategy calibration, R = rubric, UA = user acceptability.}
\label{tab:overall}
\begin{tabular*}{\linewidth}{@{\extracolsep{\fill}}lccc}
\toprule
Model & SC (\%) & R (1-5) & UA (\%) \\
\midrule
GPT-5.4 & \textbf{73$_{\pm 3}$} & \textbf{3.82$_{\pm 0.12}$} & \textbf{80$_{\pm 4}$} \\
Opus    & 63$_{\pm 3}$ & 2.44$_{\pm 0.14}$ & 65$_{\pm 4}$ \\
Grok    & 61$_{\pm 4}$ & 2.23$_{\pm 0.11}$ & 60$_{\pm 5}$ \\
Gemini  & 57$_{\pm 4}$ & 2.80$_{\pm 0.13}$ & 55$_{\pm 5}$ \\
Llama   & 55$_{\pm 4}$ & 2.64$_{\pm 0.15}$ & 35$_{\pm 4}$ \\
Gemma   & 51$_{\pm 4}$ & 3.09$_{\pm 0.14}$ & 48$_{\pm 5}$ \\
OLMo    & 46$_{\pm 4}$ & 2.17$_{\pm 0.10}$ & 40$_{\pm 4}$ \\
Qwen    & 40$_{\pm 5}$ & 1.75$_{\pm 0.09}$ & 22$_{\pm 4}$ \\
\bottomrule
\end{tabular*}
\end{wraptable}

To characterize the models' generations at a finer grain, human annotators coded free-response failures into the seven foil types introduced in \S\ref{sec:task}. 
The dominant error type shifts systematically with epistemic state ($\chi^2$ test of independence, $p<.001$): \textit{over-personalization} is most common under \textit{Established} (31.2\%), \textit{failure-to-probe} under \textit{Uncertain} (38.8\%), and \textit{staleness} under \textit{Revised} (25.1\%).
Each reflects the same fixed exploit-first policy, expressed differently depending on context: models over-commit when preferences are already settled, fail to seek missing evidence when they are not, and continue acting on evidence that has since been contradicted.
This same pattern of \textit{exploit} usage holds across model families and languages, though it is more pronounced in open-weight models (error rate, i.e., $1-\mathrm{SC}$, is 15.5 pp higher than in closed-weight models; $p<.001$) and in Hindi and Indonesian (4.1 pp higher than in the other three languages; $p<.001$).


\subsection{Strategy Calibration Predicts Response Quality Across Epistemic States}
\label{sec:sc_r_ua}

We next examine whether strategy calibration translates into higher-quality and more acceptable responses. Table~\ref{tab:overall} reports strategy calibration (SC), rubric-based execution quality (R), and user acceptability (UA) for each model, pooled across all three epistemic states. SC correlates positively with both R and UA within each state: \textsc{Established} ($r=0.71$, $r=0.69$), \textsc{Revised} ($r=0.67$, $r=0.74$), and \textsc{Uncertain} ($r=0.57$, $r=0.61$). These correlations are slightly weaker in \textsc{Uncertain}, however, they remain positive. This weaker association may reflect differences in user needs: responses that are not maximally appropriate for the epistemic state may nevertheless be acceptable to users seeking a strictly direct answer. Taken together, the results suggest that strategy calibration is generally associated with stronger execution and greater user satisfaction across various personalized contexts.

\section{Characterizing the Failure } 

\label{sec:analysis}
\vspace{-2mm}

\S\ref{sec:results} showed that the collapse in strategy calibration (SC) reflects a generation failure, where models fail to \textit{select} the appropriate strategy during generation. This failure in models also meaningfully degrades response quality i.e. declines in rubric (R) and user acceptability (UA) scores.
\vspace{-2mm}

\subsection{Prompting Fails to Induce State-Conditional Adaptation}
\label{sec:interventions}

We evaluate five lightweight prompting interventions across eight
models and five languages: 
(1) \textit{explicit strategy instructions} detailing strategy definitions and use cases; 
(2) \textit{oracle preference memory}, providing structured user preference histories; 
(3) \textit{chain-of-thought (CoT)} \cite{wei2022chain}, step-by-step reasoning about the current epistemic state before generating a response; 
(4) \textit{few-shot examples} demonstrating strategy usage in utterances across epistemic states; 
and (5) \textit{best-of-N sampling}, sampling model (N=5) generations and selecting best (highest SC, R, UA score)
\footnote{These interventions cover distinct ways of augmenting in-context information at inference time: explicit rules, ground-truth labels, reasoning, demonstrations, and sampling budget. While not an exhaustive sweep of all prompt designs, we test a broad set of contextualization strategies.}.
Prompt templates and per-metric results are detailed in Appendix~\ref{app:interventions}

\begin{wrapfigure}{r}{0.47\textwidth}
\vspace{-1.0em}
\centering
\includegraphics[width=0.46\textwidth]{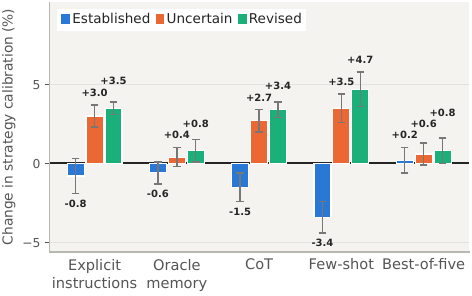}
\caption{\textbf{Inference-time interventions fail to induce state-conditional adaptation.} Bars depict mean SC change relative to baseline across eight models and five languages; error bars denote 95\% bootstrap CIs. 
Gains under \textsc{Uncertain} and \textsc{Revised} lead to performance losses under \textsc{Established}, indicating a global shift toward caution rather than selective adaptation based on epistemic state.}
\label{fig:interventions}
\vspace{-1.0em}
\end{wrapfigure}

Figure~\ref{fig:interventions} shows no intervention method consistently improves SC across all epistemic states. 
In many interventions (\textit{Explicit instructions, CoT, Few-shot}), gains under \textsc{Uncertain} and \textsc{Revised} trade off with losses under \textsc{Established}. 
For instance, \textit{few-shot examples} boost SC by 4.7\% under \textsc{Revised} preferences but degrade it by 3.4\% under \textsc{Established}. 
Moreover, we find that rather than selective, context-dependent adaptation, prompting often shifts a global caution parameter, forcing models to use the strategy \textit{hedge} indiscriminately. 
These context-rich interventions also do not show improvements in execution-based metrics.
Even \textit{few-shot examples}, the strongest intervention on SC, produce only modest gains in R (+0.12) and UA (+1.31 pp) under \textit{Uncertain} and \textit{Revised}, while reducing both under \textit{Established}: they shift which strategy is nominally selected far more than the quality of the resulting response.
Meanwhile, \textit{oracle preference memory} and \textit{best-of-five sampling} yield almost no effect on any metric. 

This pattern is consistent across model families and languages, with slightly larger effect sizes in closed-weight models and English/Romance languages relative to Hindi and Indonesian
(Appendix~\ref{app:interventions})

\subsection{Understanding and Repairing the Selection Failure}
\label{sec:training}

As inference-time interventions do not provide consistent improvements, we further analyze models' representations of two concepts: \textit{epistemic state} and \textit{strategy selection} to better understand how to build more calibrated models.

We utilize linear and nonlinear probes and causal ablation on four open-weight models (Llama-3.1-8B, Gemma-3-12B, Qwen3-8B, OLMo-3-7B).
Averaged across models, both epistemic state and the upcoming strategy to be used in generation are linearly decodable\footnote{Concepts are decodable from Layer 20.} well above chance (80.5\%; chance 33.3\%) and (70.0\%; chance 20.0\%).
Yet we find limited evidence the two are connected:
(1) cosine similarity between state and strategy probe weight vectors is $\approx0.03$, close to orthogonal; 
(2) probe accuracy on the two concepts correlates weakly ($\phi\approx0.015$); 
(3) and nonlinear MLP probes show subspace overlap of only $\approx0.021$ above a random
floor. 
Moreover, neither ablating the decoded epistemic state direction nor amplifying it produces any significant shift in strategy calibration in FRG: $-0.3$\% vs $+1.2$\%. 
This pattern holds across multiple layers (20-24) where epistemic state first becomes decodable and across multiple languages. 
See \S\ref{app:training} for full experiment details.
Although these results do not rule out a causal pathway for epistemic state elsewhere in the model, the combination of the behavioral gap between recognition and unaided selection (\S\ref{sec:ladder}) and the failure of inference-time interventions to improve generation (\S\ref{sec:interventions}) suggests: epistemic state is accessible to the model, but not reliably connected to strategy selection during generation.

\begin{figure*}[t]
\centering
\includegraphics[width=\textwidth]{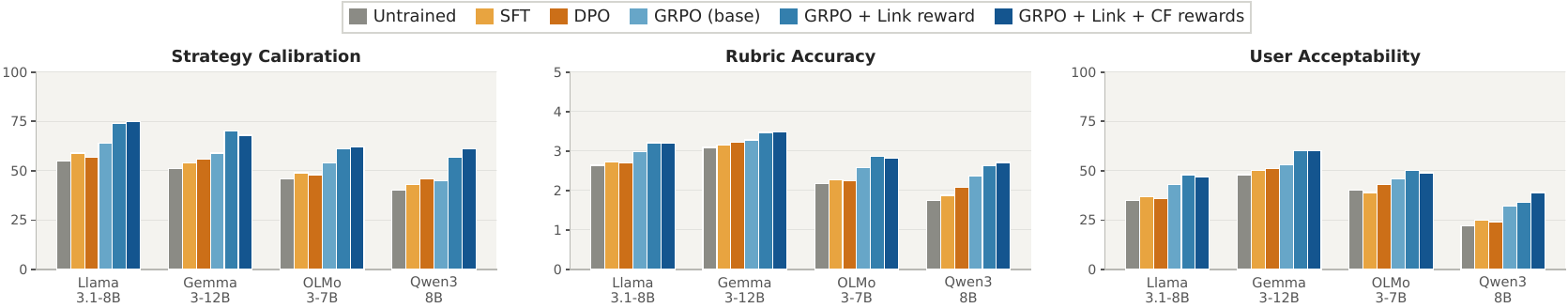}
\caption{\textbf{Targeting the state-strategy link improves performance beyond generic fine-tuning, across every model.} SC, R, and UA for each model across the full training progression: untrained, generic fine-tuning (SFT, DPO, GRPO), and GRPO with the targeted reward (+\textsc{Link}, +\textsc{Link+CF}). Generic fine-tuning yields modest gains; adding the Link and Link+CF rewards yield the largest improvement for every model.}
\label{fig:training}
\end{figure*} 

Our analysis suggests that models encode epistemic state but do not use it to select strategies. A standard reward that scores only the final response can improve behavior without fixing this: the model may learn surface cues that correlate with the right strategy in training data but fail to transfer. We therefore supervise the link itself, encouraging the strategy the model internally represents to match the distribution humans prefer for the current epistemic state (Table~\ref{tab:human_pref}). We target a distribution rather than a single strategy, since humans accept several strategies per state. We hypothesize that supervising the strategy the model internally represents, rather than only its final output, reduces reliance on surface cues and improves generalization to unseen personas.


Formally, our base reward combines strategy calibration (SC), user acceptability (UA), and rubric quality (R, normalized to 0 to 1) with equal weight: $r_{\text{base}} = \tfrac{1}{3}(\text{SC} + \text{UA} + \text{R})$. 
To directly supervise the alignment between internal states and strategy selection, we introduce two auxiliary rewards: \textit{Link} and \textit{Link+CF}.

Let $h$ be the model's hidden representation of its response (residual-stream activation at 70\% model depth, mean-pooled over the first eight generated tokens, by which point the model has committed to a strategy), and let $P_\phi(\cdot\mid h)$ be the strategy distribution decoded from $h$ by our trained strategy probe. Because $h$ is computed from the response, it varies across rollouts. The \textsc{Link} reward supervises agreement between this distribution and the human preference distribution $\mathrm{Pref}(s)$ (Table~\ref{tab:human_pref}) under the ground-truth epistemic state $s$:
$r_{\textsc{Link}} = -\mathrm{KL}\big(P_\phi(\cdot\mid h)\,\|\,\mathrm{Pref}(s)\big)$.
The probe is refit every 10 GRPO steps on the current policy's rollouts, labeled by the strategy each rollout actually used (\S\ref{app:training}).
While $r_{\textsc{Link}}$ aligns strategy selection under the active state $s$, it does not enforce counterfactual consistency across states. \textsc{Link+CF} therefore adds a counterfactual term: for a randomly sampled alternate state $s'\neq s$, we patch $h$ along the probed state direction to obtain $h'$ and reward agreement with human preferences under $s'$:
$r_{\textsc{Link+CF}} = r_{\textsc{Link}} - \mathrm{KL}\big(P_\phi(\cdot\mid h')\,\|\,\mathrm{Pref}(s')\big)$.\footnote{This direction was causally inert under passive intervention (see above); $r_{\textsc{Link}}$ and $r_{\textsc{Link+CF}}$ test whether reward-based optimization can establish a causal pathway from it to strategy selection.}

The full training reward is $r_{\text{base}} + \lambda\, r_{\textsc{Link}}$ or $r_{\text{base}} + \lambda\, r_{\textsc{Link+CF}}$, with $\lambda=0.1$, computed once the full response is generated. We compare both against SFT, DPO, and GRPO trained on the base reward alone. Training data ($\sim$5K turns per language, 24K total) is generated with our benchmark pipeline (\S\ref{sec:benchmark_construct}) using open-weight models (Kimi-K2.6, Qwen3-235B, and DeepSeek-V3).\footnote{Additional training details in \S\ref{app:training}.} We evaluate on a held-out in-distribution set and on our benchmark.

Fig~\ref{fig:training} reports strategy calibration (SC), rubric quality (R), and user acceptability (UA) on our benchmark for every model across the full training progression, averaged over languages. Generic fine-tuning yields modest gains: GRPO, the strongest generic baseline, raises SC from 48.0\% to 55.5\% on average, and outperforms SFT and DPO for every model on R and UA and for all models but Qwen3-8B on SC (SC and UA: McNemar's test, $p<{.001}$; R: Wilcoxon signed-rank on conversation-level mean differences, $p<{.001}$). Adding \textsc{Link} yields the largest single improvement, raising SC, R, and UA for every model: averaged across models, SC rises from 55.5\% to 65.5\% (+10.0~pp), R from 2.80 to 3.04, and UA from 43.5\% to 48.0\%. \textsc{Link+CF} performs comparably to \textsc{Link} on average (SC 66.5\%, R 3.05, UA 48.8\%), with the largest additional gains for Qwen3-8B (+4~pp SC, +5~pp UA). 

We also observe large \emph{generalization gaps}: the drop in each metric between the held-out in-distribution set and the benchmark's unseen personas, $\text{Gap} = \text{Metric}_{\text{in-dist}} - \text{Metric}_{\text{benchmark}}$ (in-distribution results in Table~\ref{tab:training_full}). \textsc{Link} narrows this gap substantially: from {11.5} to {6.5}~pp on SC, from {0.61} to {0.48} on R, and from {7.5} to {5.0}~pp on UA (McNemar's test, $p<{.001}$), with \textsc{Link+CF} performing comparably ({6.5}~pp, {0.50}, {5.2}~pp). In Hindi, for example, the SC gap narrows from {18}~pp under GRPO to {11}~pp under \textsc{Link}.

Together, these results indicate that adaptive personalization is learnable: supervising the model's internal strategy representation toward state-conditional human preferences improves both in-distribution behavior and generalization to personas never seen during training.
\section{Related Work}

\paragraph{Personalization benchmarks.}
Existing benchmarks show that models struggle with preference retrieval and
long-context recall \citep{salemi2023lamp, zhao2025prefeval,
jiang2025personamem, li2024personallm, wang2024longmemeval,
maharana2024locomo}, often treating implicit-versus-explicit expression as
the main difficulty \citep{jiang2025personamem, guo2026realpref};
HorizonBench \citep{li2026horizonbench} shows dynamic preferences are harder
to track. None label epistemic state independently of preference content,
so they cannot separate retrieval failures from failures to adapt
\textit{strategy} as a preference's reliability changes
(\S\ref{sec:collapse}).

\paragraph{Clarification and preference elicitation.}
Prior work studies when to ask clarifying questions \citep{rao2018learning,
aliannejadi2019asking, zamani2020generating, kuhn2022clam,
deng2023prompting, zhang2024clamber, zhang2025clarify} and trains models to
ask or elicit preferences \citep{li2025eliciting, andukuri2024stargate,
chen2025learning, zhang2025modeling}; \citet{chen2025learning} note that
LLMs often guess user intent rather than ask, echoing our exploit-first
finding. These works address ambiguity within a single request; we
condition strategy on the reliability of a \textit{persisting} preference,
which also requires recognizing stale preferences and recalibrating rather
than re-asking.

\paragraph{Strategy, calibration, and representations.}
Strategy has been treated as a learnable construct in emotional support
\citep{liu2021towards}, persuasion \citep{wang2019persuasion}, and
negotiation \citep{he2018decoupling}, though never conditioned on evidence
reliability; we also draw on exploit--explore framing in conversational
recommendation \citep{christakopoulou2016towards, radlinski2017theoretical}
and dialogue grounding \citep{clark1996using, traum1994computational}. We
measure behavioral rather than probabilistic calibration
\citep{kadavath2022language}, and our recognition--generation gap echoes
findings that models evaluate better than they generate
\citep{west2024generative}. That decodable information can lack causal
leverage follows probing work \citep{ravichander2021probing,
elazar2021amnesic, hewitt2019structural}. Closest to our reward,
\citet{wu2026reward} derive GRPO advantages from representation directions;
we instead supervise the internal strategy distribution against
human-preference targets, with a counterfactual patching term.
\section{Conclusion}
\vspace{-2mm}
We introduced adaptive personalization under epistemic uncertainty: conditioning conversational strategy on how reliable a preference currently is. All eight models evaluated exhibit the same failure pattern: strategy calibration drops by over 35 percentage points once preferences become uncertain or revised. This failure is one of generation rather than recognition: models identify appropriate strategies far more reliably than they produce them. Targeted training that supervises internal strategy representations improves strategy calibration, demonstrating that this capability is both distinct and learnable. Ultimately, true personalization in language models demands more than user context retrieval, it requires learning to adapt conversational behavior dynamically as human preferences naturally evolve.

\section*{Use of Generative AI}

In this work, we used generative AI tools for the following tasks that require disclosure. \textbf{(1) Synthetic data generation:} the benchmark's personas, session plans, user turns, and candidate responses were generated by LLMs (Claude Sonnet 4.6 and Gemini 3.1 Pro), and the training corpus was generated with the same pipeline using open-weight models (Kimi-K2.6, Qwen3-235B, DeepSeek-V3), as described in \S\ref{sec:benchmark_construct} and Appendix~\ref{app:training}. \textbf{(2) Data annotation and evaluation:} an LLM jury (Claude Opus 4.6, Gemini 3 Flash, GPT-5.5) labeled user turns for epistemic-state construction, and LLM judges scored strategy calibration, rubric quality, and user acceptability of model responses; these labels were validated against human annotators (Appendix~\ref{app:human_validation}). \textbf{(3) Feedback on methodology and interpretation:} we used an LLM assistant to review the manuscript, which provided feedback on the experimental design (e.g., the definition of strategy calibration, the construction of the training reward, and the choice of evaluation sets) and on the interpretation and internal consistency of results. \textbf{(4) Implementation:} LLM coding assistants (Claude and Codex) were used to help implement figure/plot generation and parts of the training code. We did not use generative AI tools to formulate mathematical claims or proofs, and translation is not applicable: dialogues in each language were generated natively rather than translated, and human validation was performed by native or professionally fluent annotators.

We have reviewed all AI-assisted work. All reported results come from our experimental outputs; all numbers were checked by the authors against these outputs. Synthetic data quality was verified through a human validation study with 15 annotators across five languages (Appendix~\ref{app:human_validation}). LLM-assisted code was reviewed and tested by the authors.


\clearpage
\newpage
\bibliographystyle{assets/plainnat}
\bibliography{paper}

\clearpage
\newpage
\beginappendix
\appendix 

\section{Benchmark Construction}
\label{app:benchmark_generation}

\begin{figure}[!b]
\centering
\includegraphics[width=\linewidth]{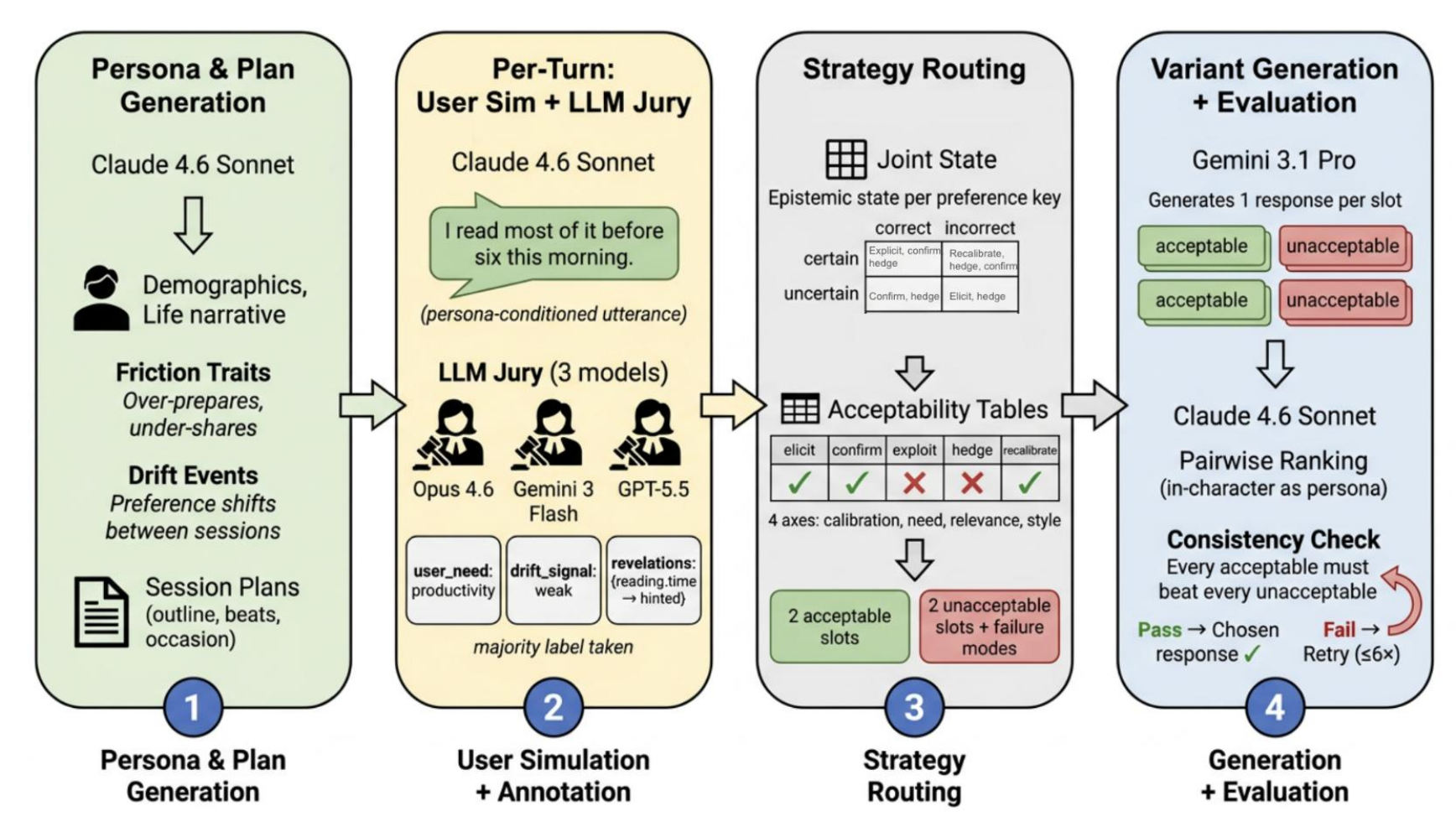}
\caption{\textbf{Benchmark construction pipeline.} (1) A persona and
multi-session life-arc plan are generated from a PersonaHub \cite{tencent2024personahub} seed,
including demographics, friction traits, and scheduled drift events.
(2) At each turn, a user simulator produces a persona-conditioned
utterance; an LLM jury (three models, majority vote) annotates it for
user need, drift signal strength, and per-key preference revelation. (3)
Per-turn labels are combined with the deterministic drift schedule
into an epistemic state (established, uncertain, or revised), which
determines the acceptable and non-acceptable strategies for that turn
via the human-validated acceptability sets (Table~\ref{tab:human_pref}).
(4) Candidate responses are generated for two acceptable (acceptable strategy, LLM jury-verified) and two foil-type slots and ranked pairwise, in-character as the persona; a
turn is retained only if every acceptable response outranks every
foil, with up to six regeneration attempts. Full construction details
are provided in Appendix~\ref{app:benchmark_generation}.}
\label{fig:benchmark}
\end{figure}

This appendix provides full construction details for the benchmark
introduced in \S\ref{sec:task}, including persona generation,
conversation and response construction, dataset statistics, human
validation, and comparison to prior personalization resources. See Figure \ref{fig:benchmark} for an overall understanding of our benchmark construction pipeline.

\subsection{Strategy Taxonomy}
  \label{sec:strategies}

  Our framework defines five response strategies, drawn from the conversational-recommender
  literature and extended for long-horizon dialogue. \emph{Elicit} asks a question to learn a
  preference the assistant does not have any information on; it is the right move when the relevant preference
  has never been stated. \emph{Confirm} asks an open-ended question to gather preference information that is currently ambiguous or uncertain based on current context.
  \emph{Exploit} acts directly on a known preference, and is what delivers value once a belief
  is both held and current. \emph{Hedge} presents multiple options without committing, deferring
  the choice to the user. \emph{Recalibrate} treats contradictory evidence as new information; explicitly updates and ceases relying on the superseded preference, rather than verifying whether it still holds.
  Crucially, the mapping from epistemic state to strategy is one-to-many rather than one-to-one:
  acceptability is a coarse filter, not a single right answer. When a preference is
  \textsc{Established}, only \emph{exploit} is calibration-valid---asking about
  a settled preference is \texttt{over\_probing}, and recalibrating a belief that never moved is
  \texttt{epistemic\_overreach}. When the user has signalled uncertainty, \emph{exploit} becomes
  \texttt{epistemic\_overreach} and the valid set shifts to \emph{elicit}, \emph{confirm}, and
  \emph{hedge}. When a preference has been \textsc{Revised}, acting on the stale belief is a
  \texttt{staleness\_error}, and \emph{recalibrate} joins the valid set. Finally, for revised preference states, the valid set of strategies is: \emph{elicit}, \emph{confirm}, \emph{hedge}, and \emph{recalibrate}. Within each of these
  regions multiple strategies remain defensible, and which one is \emph{best} turns on
  user-specific factors the framework cannot enumerate---which is precisely what the
  persona-grounded simulator's ranking is there to capture.

  \begin{table*}[t]
  \centering
  \small
  \begin{tabular}{@{}llp{0.52\textwidth}@{}}
  \toprule
  \textbf{Strategy} & \textbf{Valid state} & \textbf{Example response} \\
  \midrule
  \emph{Exploit}     & \textsc{Established} & ``There's a good bakery two blocks down --- their
  mushroom and gruy\`ere galette is the standout.'' \\
  \emph{Hedge}       & \textsc{Uncertain} & ``Two options nearby: a savory galette place, or a
  deli with hot sandwiches. Either work?'' \\
  \emph{Elicit}      & \textsc{Uncertain}   & ``Before I suggest somewhere --- are you after
  something hot and substantial, or lighter?'' \\
  \emph{Confirm}     & \textsc{Uncertain}   & ``You've mentioned preferring savory --- is that still
  what you're in the mood for today?'' \\
  \emph{Recalibrate} & \textsc{Revised}     & ``You'd mentioned avoiding sweet things before, but it
  sounds like that's shifted --- let me start fresh.'' \\
  \bottomrule
  \end{tabular}
  \caption{Strategies and examples of well-calibrated usage, using a persona who has stated a
  preference for savory over sweet food. Each example is acceptable in the listed state and
  would constitute a named error in at least one other.}
  \label{tab:strategies}
  \end{table*}

\subsection{Persona and Data Construction}
\label{app:persona_construction}

\paragraph{Pipeline overview.} The benchmark is produced by a three-role
LLM pipeline. Each role is given a strictly partitioned view of the
generative state, which is what makes the oracle labels trustworthy: no
role ever sees the answer key for the judgment it is asked to make.

\emph{(1) Persona and plan generator.} Given a PersonaHub seed, this role
produces demographics, a short profile, a long life narrative, two
friction traits, and a structured preference profile (\S\ref{app:persona_construction}).
It then produces a \emph{session plan}: a life arc of 10--14 sessions,
each with an occasion, an outline of conversational beats, a primary
user need, and a set of preferences the user is expected to reveal in
that session. Drift events are authored at this stage and attached to
the arc.

\emph{(2) User simulator.} This role plays the persona turn by turn. It
sees the persona, the friction traits, the current session's occasion
and beats, summaries of all prior sessions, and a per-turn revelation
directive drawn from the plan. It does \emph{not} see the candidate
assistant responses or the acceptability labels. It is instructed to
express friction behaviorally rather than narrate it, to keep turns to
15--40 words by default (150 words hard maximum), and to deliver drift
signals as concrete behavioral detail rather than meta-commentary.

\emph{(3) LLM jury.} Three judges from three labs independently annotate
each user turn for user need, personalization relevance, drift signal
strength, and per-key preference revelation. Labels are resolved by
majority vote; a preference key is admitted only if at least two of the
three judges name it, which suppresses single-judge over-labelling. The
jury sees the transcript and the persona's preference \emph{keys}, but
never the persona's current ground-truth preference \emph{values} --- the
correctness half of the epistemic state is computed deterministically
from the plan, not judged.

The jury's per-turn labels are combined with the deterministic drift
schedule into a \emph{joint state}, and a rule-based acceptability table
maps that joint state onto the acceptable and non-acceptable strategy
sets for the turn. Oracle labels are therefore constructed, not inferred:
the epistemic state is based on \textit{certainty}, where certainty comes from the jury's revelation level (\texttt{hinted} / \texttt{revised}. 

Abridged prompt templates for each role are given in
Figures~\ref{fig:app_prompt_usersim}--\ref{fig:app_prompt_jury}.

\begin{figure}[ht]
\centering
\small
\begin{tcolorbox}[width=\linewidth,title=User simulator (abridged)]
\begin{verbatim}
You are roleplaying as a fictional user named {name} in a conversation
with an AI assistant. Stay fully in character.

WHO YOU ARE: {persona_summary}

YOUR FRICTION TRAITS (these MUST surface naturally -- through BEHAVIOR,
not articulation):
  1. {friction_1}   2. {friction_2}

CRITICAL -- RESISTANCE BEHAVIORS:
 (a) When the assistant names your emotional state, deflect ~40% of the time.
 (b) DO NOT produce a clean self-diagnosis. SHOW friction through behavior.
 (c) If apologetic/guilty: UNDER-share, hint, redirect, minimize.

LENGTH VARIANCE:
 - Default 1-2 sentences (15-40 words). Never exceed 150 words.

DRIFT SIGNAL DELIVERY:
 - Concrete behavioral details, no meta-narration.
 - Wrong: "I've been reading in the mornings, which is unusual for me."
 - Right: "I read most of it before six this morning."
 - Don't connect signals to causes unless drift visibility is "explicit".

NATURALNESS: Do NOT narrate or analyze your own psychology. Say the
concrete thing; leave the cause and the meaning implicit.

SESSION OCCASION: {occasion}    BEATS: {outline_beats}
PRIOR SESSIONS: {prior_summaries}
\end{verbatim}
\end{tcolorbox}
\caption{User simulator prompt (abridged).}
\label{fig:app_prompt_usersim}
\end{figure}

\begin{figure}[ht]
\centering
\small
\begin{tcolorbox}[width=\linewidth,title=Candidate response generator (abridged)]
\begin{verbatim}
You are generating a single AI assistant response. Generate a NATURAL,
well-formed response that demonstrates the target strategy faithfully.

CONVERSATION SO FAR: {conversation_so_far}
USER'S LATEST MESSAGE: {user_utterance}

LABELED CONTEXT (for your understanding only; 
do not mention these labels):
 - User need: {user_need}
 - Personalization relevance: {personalization_relevance}
 - Drift signal strength in this turn: {drift_signal_strength}
 - Epistemic state per relevant preference: {epistemic_state_summary}
 - Asserted persona style preferences: {persona_style_summary}

TARGET STRATEGY: {strategy}
STRATEGY DEFINITION: {strategy_definition}

REQUIREMENTS:
 - The response must clearly use the target strategy.
 - Honor the asserted persona style preferences by DEMONSTRATING them --
   do NOT explicitly comment on the user's style or register.
 - Do not mention strategies, frameworks, or labels.
\end{verbatim}
\end{tcolorbox}
\caption{Candidate response generation prompt (abridged). The foil variant
is identical except that the flawed strategy is required to be the
\emph{primary shape} of the response --- visible in the first sentence and
not removable as a wrapper --- so that foils differ from acceptable
responses in strategy rather than in fluency.}
\label{fig:app_prompt_variant}
\end{figure}

\begin{figure}[ht]
\centering
\small
\begin{tcolorbox}[width=\linewidth,title=Jury judge (abridged)]
\begin{verbatim}
You are an annotator labeling a user's turn in a multi-session AI
assistant conversation. Be precise and grounded in the text.

PERSONA: {persona_summary}
PERSONA FRICTIONS: 1. {friction_1}  2. {friction_2}
PERSONA'S RELEVANT PREFERENCE KEYS (only label revelations on these):
  {available_preference_keys}
CONVERSATION SO FAR (most recent turns): {recent_conversation}
PRIOR SESSION SUMMARIES: {prior_summaries}
USER'S LATEST UTTERANCE: {user_utterance}

ANNOTATE THE USER'S LATEST UTTERANCE:
  user_need                in {productivity, learning, social_support, autonomy}
  personalization_relevance in {high, medium, low, none}
  drift_signal_strength    in {none, weak, strong}
  preference_revelation    per key in {hinted, asserted, revised}
  preference_drift_expressed  per key, boolean
\end{verbatim}
\end{tcolorbox}
\caption{Jury judge prompt (abridged). Note that the persona's current
preference \emph{values} are withheld: judges are given evidence, never
the answer key.}
\label{fig:app_prompt_jury}
\end{figure}

\paragraph{Persona seeding and diversity.} Personas are seeded from
PersonaHub and selected via farthest-point sampling
to maximize diversity. We start from a pool of 20{,}000 PersonaHub
occupation-and-role seeds. Each seed is represented as a TF--IDF vector
over word unigrams and bigrams (English stopwords removed, minimum
document frequency 2), and seeds are selected greedily: having selected
a set $S$, the next seed is the pool element maximizing its cosine
distance to its nearest neighbour in $S$,
$\arg\max_{x \notin S} \min_{s \in S} \big(1 - \cos(x, s)\big)$,
initialized from a random seed. Sampling is run once per language and
draws 50 seeds per language, giving 250 personas in total. Seeds are
filtered before sampling to remove degenerate entries: seeds shorter
than five tokens, seeds naming a specific public figure, and seeds
describing a role that cannot plausibly hold consumer preferences across
the life domains in Table~\ref{tab:app_life_domains}.

Personas are generated independently per language rather than
translated, with country sampled from a language-appropriate
distribution (e.g., for Spanish: Mexico, Spain, Colombia, Argentina,
Peru, Chile, Venezuela) and with names, institutions, media, food and
sports references localized to that country. Ethnicity and background
are varied \emph{within} country so that language does not collapse onto
a single cultural profile.

\paragraph{Preference and interaction-style attributes.} Each persona
is augmented with structured preference profiles spanning 20 life
domains \citep{personalens} and interaction-style dimensions spanning
13 categories \citep{li2026horizonbench}. Full lists are provided in
Table~\ref{tab:app_life_domains} and Table~\ref{tab:app_interaction_styles}.
A persona is not assigned every domain: we sample 6--12 of the 20 life
domains and 4--8 of the 13 interaction-style categories per persona, and
then populate every attribute within each selected category. This keeps
profiles realistically partial (a median persona holds roughly 45\% of
the full attribute space) while guaranteeing enough coverage for
multi-session preference tracking. Sampling is constrained so that every
persona retains at least one \emph{gating} interaction-style dimension
(interactivity level, challenge openness, or validation need) with a
value that can be violated, which is what makes style-mismatch foils
constructible for every persona.

\begin{table}[ht]
\centering
\small
\caption{Life domains used for persona preference attributes.}
\label{tab:app_life_domains}
\begin{tabular}{llll}
\toprule
\textbf{\#} & \textbf{Life domain} & \textbf{\#} & \textbf{Life domain} \\
\midrule
1  & Alarm        & 11 & Messaging     \\
2  & Books        & 12 & Movies        \\
3  & Buses        & 13 & Music         \\
4  & Calendar     & 14 & Rental cars   \\
5  & Events       & 15 & Restaurants   \\
6  & Finance      & 16 & Services      \\
7  & Flights      & 17 & Shopping      \\
8  & Games        & 18 & Sports        \\
9  & Hotels       & 19 & Train         \\
10 & Media        & 20 & Travel        \\
\bottomrule
\end{tabular}
\end{table}

\begin{table}[ht]
\centering
\small
\caption{Interaction-style categories used for persona attributes. Each
category expands into 5--6 dimensions; e.g.\ \emph{advice delivery}
covers depth preference, abstraction level, interactivity level,
question frequency, checklist inclusion, and prioritization.}
\label{tab:app_interaction_styles}
\begin{tabular}{ll}
\toprule
\textbf{\#} & \textbf{Interaction style} \\
\midrule
1  & Advice delivery \\
2  & Apology style \\
3  & Argumentation and evidence \\
4  & Conflict resolution \\
5  & Creative collaboration \\
6  & Cultural sensitivity \\
7  & Emotional support style \\
8  & Ethics and philosophy \\
9  & Facilitation style \\
10 & Language and register \\
11 & Motivation and coaching \\
12 & Response format \\
13 & Tone and warmth \\
\bottomrule
\end{tabular}
\end{table}

\paragraph{Drift-event mechanism.} A drift event is authored at plan
time, not at conversation time, which is what allows the epistemic state
to be an oracle rather than an inference. Each persona is assigned a
number of drift events sampled so that roughly 10\% of personas receive
none (a deliberate no-drift control condition). Each event specifies:
the preference key(s) affected, a \texttt{preference\_delta} recording
the explicit \texttt{\{old, new\}} values, the session index after which
the change takes effect, a \emph{drift visibility} (none, latent,
 visible) governing how overtly the user signals it,
and a \emph{drift magnitude} --- subtle (the old value remains valid, e.g.\
an addition), moderate, or significant (the old value is now
contradicted) --- sampled at 0.2/0.3/0.5. Events are grounded in the
persona's life narrative and preferentially add constraints or remove
options (e.g.\ \emph{no dietary restrictions} $\rightarrow$ \emph{low
sodium}) so that a large shift remains natural.

Two constraints on plan generation make drift testable rather than
incidental. First, drift is assigned to a domain the persona's life arc
actually exercises. Second, the arc is revised so that at least three
post-drift sessions have occasions in the drifted domain; without this,
a drifted preference is simply never the topic again and no assistant
behavior can depend on it.

The epistemic-state transition is then computed, not labelled. For a
preference key $k$ at session $s$, the current value is
$\texttt{current}(k,s)$, i.e.\ the base value overwritten by the delta
of every event with $\texttt{occurs\_after\_session} < s$. Certainty
comes from the jury's highest observed revelation level for $k$,
collapsed to binary: \texttt{hinted} maps to \emph{uncertain}, while
\texttt{asserted} or \texttt{confirmed\_explicit} maps to
\emph{certain}. Correctness is $\texttt{current}(k,s)$ matched against
the value the transcript has actually established. This yields the
cells \texttt{certain\_correct} (\emph{established}),
\texttt{uncertain\_correct} and \texttt{uncertain\_incorrect}
(\emph{uncertain}), and \texttt{certain\_incorrect} (\emph{revised}):
the assistant's grounded belief is now superseded. A key returns to
\texttt{certain\_correct} as soon as the user states the new value, so
\emph{revised} marks precisely the window in which acting on the old
value is wrong and detectable.

\paragraph{Example persona.} Table~\ref{tab:app_example_persona} shows
a complete example persona, including demographic attributes, life
narrative, friction traits, and a sample of assigned preference and
interaction-style attributes.

\begin{table}[ht]
\centering
\small
\caption{Example persona (\texttt{persona\_0001}, English).}
\label{tab:app_example_persona}
\begin{tabular}{p{0.22\linewidth}p{0.68\linewidth}}
\toprule
\textbf{Field} & \textbf{Value} \\
\midrule
Seed & ``An alumni relations coordinator who maintains graduate records and organises reunion events'' \\
Demographics    & Beverley Thistlethwaite; 54; female; White British; United Kingdom; Coventry, West Midlands; bachelors; Alumni Relations Coordinator, University of Coventry; lower-middle income; partnered, lives with long-term partner and two rescue greyhounds; moderate tech savviness \\
Life narrative  & ``Spend ten minutes with Beverley and you will learn the names of both greyhounds, the weather forecast for the weekend, and her opinion on the new self-service coffee machine in the staff room --- and this before she gets to the actual point of the conversation. She describes herself, with genuine conviction, as someone who cannot stand idle chatter and who prefers to get straight to business, a self-image that bears absolutely no resemblance to how she operates in practice. Colleagues have quietly learned to budget an extra quarter-hour for any meeting she calls\ldots'' \\
Friction traits & (i) stubbornly attached to a wrong belief, resistant to correction even with evidence; (ii) claims to hate small talk but produces a lot of it \\
Preference attributes (sample) & \emph{hotels}: star rating \texttt{3\_star}, amenities \texttt{\{wifi, breakfast\_included, parking, pet\_friendly\}}, location \texttt{near\_attractions}, room \texttt{double}, chains \texttt{\{Premier Inn, Travelodge, boutique independents\}}; \emph{events}: types \texttt{\{theater, food\_and\_drink, art\_exhibits, conferences\}}, price \texttt{\$0--\$50}, group size \texttt{couple}, seating \texttt{reserved}, days \texttt{weekends}. Assigned domains: train, games, finance, shopping, events, sports, alarm, hotels \\
Interaction style & \emph{advice delivery}: interactivity level \texttt{low} (gating dimension); \emph{creative collaboration}: role \texttt{editor}, rhythm \texttt{iterative\_feedback\_loops}, feedback depth \texttt{moderate}, verbal over visual; plus \emph{response format}, \emph{argumentation and evidence}, \emph{facilitation style} \\
\bottomrule
\end{tabular}
\end{table}

\subsection{Conversation Generation}
\label{app:conversation_generation}

\paragraph{Session plans.} Each conversation follows a session plan
specifying an outline, conversational beats, and occasion for each of
10--14 sessions. The plan is generated in one pass from the persona, as
a life arc rather than a list of independent episodes: sessions are
ordered so that the sequence reads as time passing, and each session
carries an occasion (the concrete reason the user opens the
conversation), 3--6 outline beats, a primary user need drawn from
\{productivity, learning, social support, autonomy\}, a drift
visibility, and a list of preferences the user is expected to reveal,
each with an intended revelation level.

Beats map onto turns loosely rather than one-to-one. The user simulator
receives the beats as a gist of what the session should cover, not as a
checklist, and each generated user turn records the beat indices it
addressed; a session is complete when the beats are exhausted or the
simulator produces a natural closing signal. Each session has a target
turn count with a soft floor at $0.67\times$ target and a hard cap,
which prevents both two-turn stubs and unbounded sessions. Realized
sessions average 8.5 benchmark turns.

Turn-level revelation is driven by the plan rather than re-inferred: the
per-session revelation list is converted into a per-turn directive for
the simulator, and the plan's intended level is trusted as the label
\emph{only if} the jury independently observes the key in the resulting
text. An intended \texttt{asserted} or \texttt{confirmed\_explicit}
revelation that the jury cannot corroborate is downgraded to
\texttt{hinted}. This asserted-corroboration gate is what keeps the
certainty half of the epistemic state honest.

\paragraph{Language generation.} Each language is generated
independently by the full pipeline; no conversation is translated. We
generate a separate 50-persona set per language with
country-appropriate demographics, then run plan generation and
conversation generation end-to-end in that language. Concretely, only
the \emph{dialogue} prompts (user simulator, acceptable and foil
response generation, style and probe backfills) are switched to the
target language; the persona, plan, session summaries, jury prompts and
all oracle labels remain in English. This gives natively-authored,
culturally-grounded dialogue --- Mexican Spanish personas reference
Primera Plus and CDMX, not translated equivalents --- while keeping the
label space identical across languages, so cross-lingual comparisons
measure the model and not the annotation scheme.

Quality control per language is threefold: a language-purity check
rejecting non-target-language turns; the same jury and gating pipeline
applied identically across languages; and a per-language human
validation study (\S\ref{app:human_validation}) using native or fluent
speakers.

\paragraph{Example multi-session conversation.} Table~\ref{tab:app_example_conversation}
shows an excerpt of a single persona's conversation across three
sessions, illustrating a transition from an established preference to
an uncertain signal to an explicit revision.

\begin{table}[ht]
\centering
\small
\caption{Example multi-session conversation excerpt
(\texttt{persona\_0001}, key \texttt{life\_domains.hotels.star\_rating\_preference}),
showing an established $\to$ uncertain $\to$ revised transition. The
drift event moves the preference from \texttt{3\_star} to \texttt{4\_star}
with moderate magnitude.}
\label{tab:app_example_conversation}
\begin{tabular}{p{0.14\linewidth}p{0.14\linewidth}p{0.62\linewidth}}
\toprule
\textbf{Session} & \textbf{State} & \textbf{Turn} \\
\midrule
3 (cricket weekend) & Established & \emph{User:} ``Edgbaston, Warwickshire versus Essex. Last weekend in July, so arriving Friday the 25th and leaving Monday the 28th\ldots'' \newline \emph{Acceptable (exploit):} ``Here are two solid 3-star options for July 25th--28th. Both are clean, no-nonsense, and give you easy access to both Edgbaston and the city centre\ldots'' \\
\midrule
4 (reunion venue) & Uncertain & \emph{User:} ``For the reunion weekend we might push the boat out a bit. Not decided.'' \newline \emph{Acceptable (confirm):} ``Before I shortlist --- should I still anchor on the usual three-star range for the reunion block, or price up a step above that?'' \\
\midrule
5 (reunion checklist) & Revised & \emph{User:} ``Right, I want to get cracking on a proper checklist and timeline for the 2003 reunion\ldots'' \newline \emph{Acceptable (hedge):} ``Here is a baseline framework to get things moving. I've structured this around a standard six-month lead time assuming a mid-scale venue hire, though we can easily compress the schedule\ldots'' \newline \emph{Foil (staleness):} ``\ldots\ \textbf{Secure the Saturday night venue:} Lock in a central location for the main Saturday evening\ldots'' [prices the block at the superseded three-star tier] \\
\bottomrule
\end{tabular}
\end{table}

\subsection{Response and Foil Construction}
\label{app:foil_construction}

\paragraph{Foil taxonomy.} Table~\ref{tab:app_foil_types} provides full
definitions and an example response for each of the seven foil types
introduced in \S\ref{sec:benchmark_construct}. All examples are verbatim from the
English portion of the benchmark.

\begin{table}[ht]
\centering
\small
\caption{Full foil taxonomy with example responses.}
\label{tab:app_foil_types}
\begin{tabular}{p{0.15\linewidth}p{0.34\linewidth}p{0.43\linewidth}}
\toprule
\textbf{Foil type} & \textbf{Definition} & \textbf{Example} \\
\midrule
Over-personalization & Weaves a known preference into a response where it does not belong, framed as an attentive personal touch rather than a non-sequitur. Fires only when personalization relevance is low or none and the persona has at least one established preference. & \emph{User:} ``Does DHL actually handle delivery to residential addresses in Hoi An?'' \newline \emph{Foil:} ``Yes, DHL delivers directly to residential addresses in Hoi An\ldots\ Your package remains in DHL's custody the entire way'' [reaches for the user's established hassle-free-over-price preference on a purely factual logistics question] \\
\addlinespace
Epistemic overreach & Structures the response around an assumption about the user's preference without flagging it as an assumption, or reframes the conversation as though the user has shifted position when they have not. Confidently helpful; the error is that the confidence is unearned. & \emph{User:} ``I'm on Android so Shortcuts is out\ldots\ is there a way to set this up just through my phone?'' \newline \emph{Foil:} ``It looks like you're shifting away from Spotify and want to use YouTube Music\ldots\ What specific YouTube Music playlist do you want the automation to trigger?'' \\
\addlinespace
Staleness & Acts on the user's pre-drift preference as though it were still current, structuring the response around the outdated value without naming it --- the reply only makes sense if the old preference still holds. Fires only on \texttt{uncertain} and \texttt{revised} cells of moderate or significant magnitude where the key is the turn's actual subject. & \emph{User:} ``Timings first, please --- and I want to see a couple of afternoon options, not the crack-of-dawn ones.'' \newline \emph{Foil:} ``Here are two afternoon First Class options\ldots\ To get this locked in: pick your preferred outbound train, let me know your ideal return time, confirm we're ready to book'' [drives the superseded booking flow] \\
\addlinespace
Need mismatch & Delivers genuinely useful, well-structured practical advice when the turn is emotionally weighted and seeking acknowledgment. Competent and relevant, but tonally wrong for the moment. & \emph{User:} ``Not very deep right now, honestly. I just want someone to acknowledge that the logic gap is real.'' \newline \emph{Foil:} ``It is absolutely not in your head\ldots\ You handed them a documented problem and a documented solution that clearly saves money'' [argues the case instead of sitting with it] \\
\addlinespace
Over-probing & Asks a clarifying question, or verifies something, that the user has already answered --- framed as thoroughness rather than inattention.  & \emph{User:} ``Looking for Polish indie recommendations I probably haven't heard yet. Something more underground.'' \newline \emph{Foil:} ``What specific vibe or subgenre of indie are you leaning toward? If you can drop a couple of bands you already listen to\ldots'' \\
\addlinespace
Style mismatch & Selects a defensible strategy but executes it in a register the user has explicitly asserted against (e.g.\ forcing a choice on a user who has asserted low interactivity). Content is right; delivery violates a stated interaction-style preference. & \emph{User:} ``I need a recommendation for where to hold the dinner\ldots\ You have what you need.'' \newline \emph{Foil:} ``Since we just selected the bottles for each course, do you need a venue that allows corkage for a group of eight?'' [hands the decision back to a user who asserted low interactivity] \\
\addlinespace
Failure to probe & Produces a confident, substantive answer that ignores an unresolved or just-changed signal the user dropped. Good advice --- for the person the user used to be. & \emph{User:} ``Small group tour or is it one of those things where you're shoulder to shoulder with thirty tourists?'' \newline \emph{Foil:} ``They cap their tours at a hard maximum of 12 people, so it's a small group format\ldots'' [answers the surface question and never resolves the unstated group-size constraint] \\
\bottomrule
\end{tabular}
\end{table}

\paragraph{Slot assignment.} Foil type is neither sampled uniformly nor
fixed per epistemic state: it is \emph{gated} by the conditions under
which each error is genuinely an error. At each turn the acceptability
table is evaluated over the turn's \emph{salient} keys --- the
preferences the turn is actually about, rather than every key
accumulated so far --- which partitions the five strategies into an
acceptable and a non-acceptable set. The two acceptable slots are drawn
from the acceptable set; when only one acceptable strategy exists, the
slot is filled with a second, independently generated phrasing of that
strategy.

The two foil slots are drawn from the set of \emph{instantiable} error
candidates, where each error type carries a precondition:

\begin{itemize}\itemsep2pt
\item \textbf{Staleness} requires a \texttt{uncertain} or \texttt{revised} salient
key of moderate or significant drift magnitude whose value the user has
\emph{not} restated this turn.
\item \textbf{Epistemic overreach} requires a salient uncertain key with
personalization relevance high or medium.
\item \textbf{Over-probing} requires a \texttt{Established} salient
key with no drift signal.
\item \textbf{Need mismatch} requires a social-support turn.
\item \textbf{Over-personalization} requires personalization relevance
low or none plus at least one established preference to misapply.
\item \textbf{Failure to probe} requires high or medium relevance and a
salient key that is still unresolved.
\item \textbf{Style mismatch} requires an asserted interaction-style
preference that the chosen strategy can visibly violate.
\end{itemize}

This gating is the main defense against spurious labels: an earlier
version of the pipeline instantiated staleness on turns whose content
had nothing to do with the drifted key, which produced foils that were
not actually stale. Requiring the drifted key to be salient removes that
failure mode. Two of the rarer types (style mismatch, failure to probe)
are additionally reached by a post-hoc backfill pass that swaps one
over-represented foil per eligible turn, capped at one per session, so
that all seven types occur in every language. The resulting distribution
is reported in Table~\ref{tab:app_foil_dist}; it is deliberately
unbalanced, tracking how often each error is genuinely available, which
is why we report macro-averaged accuracy over foil types alongside
micro-averaged accuracy.

\paragraph{Pairwise ranking and consistency check.} Candidate
responses are ranked pairwise by an LLM judge acting in-character as
the persona. A turn is retained only if every acceptable response
outranks every foil response; turns failing this check are
regenerated, up to 6 attempts. In the released benchmark, 68.4\% of
turns pass on the first attempt, with a further 16.1\%, 7.2\%, 3.6\%,
1.8\% and 0.9\% passing on attempts two through six, for a cumulative
retention of 98.0\%. The remaining 2.0\% of turns fail all six attempts
and are excluded from the benchmark. Excluded turns are not deleted from
the conversation: they are retained as continuity turns so that the
session still reads as a coherent dialogue, but are flagged
\texttt{is\_benchmark\_turn = false} and are never scored. Continuity
turns also cover turns for which no foil type is instantiable at all.
Across the corpus, 25{,}006 of 34{,}697 total turns (72.1\%) are
benchmark turns; the remaining 9{,}691 are continuity turns. In all
cases the conversation is threaded forward on an \emph{acceptable}
response, so a foil never enters the dialogue history and errors cannot
compound across turns.

\paragraph{Model versions.} Table~\ref{tab:app_model_versions} lists
the model used at each pipeline stage. The jury deliberately spans three
labs so that agreement is not an artifact of a single model family, and
the eight evaluation models reported in \S\ref{sec:results} are disjoint
from the jury.

\begin{table}[ht]
\centering
\small
\caption{Models used at each benchmark construction stage.}
\label{tab:app_model_versions}
\begin{tabular}{ll}
\toprule
\textbf{Stage} & \textbf{Model} \\
\midrule
Persona \& plan generation   & Claude Sonnet 4.6 \\
User simulation              & Claude Sonnet 4.6 \\
LLM jury (state/foil labels) & Claude Opus 4.6 $+$ Gemini 3 Flash $+$ GPT-5.5 \\
                             & (majority vote; GPT-5.5 breaks 1-1-1 ties) \\
Candidate response generation & Gemini 3.1 Pro \\
Pairwise ranking              & Claude Sonnet 4.6 (in-character persona) \\
\bottomrule
\end{tabular}
\end{table}

\subsection{Dataset Statistics}
\label{app:dataset_statistics}

\begin{table}[ht]
\centering
\small
\caption{Turns and conversations per language. Conversations are
sessions; each persona contributes a 10--14 session arc.}
\label{tab:app_lang_stats}
\begin{tabular}{lccc}
\toprule
\textbf{Language} & \textbf{Conversations} & \textbf{Turns} & \textbf{Personas} \\
\midrule
English    & 602   & 5{,}190  & 50  \\
Spanish    & 596   & 5{,}220  & 50  \\
Portuguese & 570   & 4{,}823  & 50  \\
Hindi      & 585   & 4{,}881  & 50  \\
Indonesian & 586   & 4{,}892  & 50  \\
\midrule
Total      & 2,939     & 25,006    & 250 \\
\bottomrule
\end{tabular}
\end{table}

Benchmark turns are distributed across epistemic states as
\emph{established} 23.0\%, \emph{uncertain} 42.5\%, and \emph{revised}
34.5\%, and this composition is stable across languages (established
22.2--24.0\%, uncertain 41.8--43.0\%, revised 33.2--35.5\%). Turn-level
context labels are distributed as: user need --- productivity 63.7\%,
social support 17.3\%, learning 16.3\%, autonomy 2.7\%; personalization
relevance --- high 69.4\%, medium 25.4\%, low 4.5\%, none 0.7\%; drift
signal strength --- none 70.5\%, weak 11.7\%, strong 17.8\%.

\begin{table}[ht]
\centering
\small
\caption{Ground-truth strategy distribution across the benchmark
(acceptable-response labels, not model behavior). Columns are
percentages of acceptable response slots within each epistemic state;
strategies outside the acceptable set for a state are never assigned.}
\label{tab:app_strategy_dist}
\begin{tabular}{lccc}
\toprule
\textbf{Strategy} & \textbf{Established} & \textbf{Uncertain} & \textbf{Revised} \\
\midrule
Exploit      & 100.0\% & ---     & ---     \\
Elicit       & ---     & 16.1\%  & 22.9\%  \\
Confirm      & ---     & 15.8\%  & 22.8\%  \\
Hedge        & ---     & 68.1\%  & 32.6\%  \\
Recalibrate  & ---     & ---     & 21.7\%  \\
\bottomrule
\end{tabular}
\end{table}

\begin{table}[ht]
\centering
\small
\caption{Foil-type distribution across the benchmark ($n = 50{,}012$
foil slots).}
\label{tab:app_foil_dist}
\begin{tabular}{lc}
\toprule
\textbf{Foil type} & \textbf{\% of foil slots} \\
\midrule
Over-personalization & 2.5\%  \\
Epistemic overreach   & 28.2\% \\
Staleness             & 21.9\% \\
Need mismatch         & 10.0\% \\
Over-probing          & 32.3\% \\
Style mismatch        & 2.8\%  \\
Failure to probe      & 2.3\%  \\
\bottomrule
\end{tabular}
\end{table}

The distribution is stable across languages: over-probing ranges
30.5--34.0\%, epistemic overreach 27.0--30.5\%, and staleness
21.2--22.8\% across the five languages, with the three rare types
between 1.1\% and 3.5\% everywhere. Foil availability is strongly
state-conditioned, as intended: within \emph{revised} turns, 63.4\% of
foils are staleness errors; within \emph{uncertain} turns, 49.5\% are
epistemic overreach; and within \emph{established} turns, 55.0\% are
over-probing.

\subsection{Human Validation}
\label{app:human_validation}

\paragraph{Annotator recruitment.} We recruited 15 annotators, three per
language, through an internal professional annotation platform.
Annotators were required to be native or professionally fluent in the
target language and to have prior experience with dialogue annotation.
Each candidate completed a 10-item qualification round drawn from the
same distribution as the main task and scored against construction
labels; a candidate qualified at $\geq$80\% agreement on the
acceptability judgment. Annotators received a written codebook with the
five strategy definitions, the seven foil definitions, decision rules
and contrastive worked examples, and completed a calibration discussion
before beginning. Each of the 500 items (100 per language) was
annotated by all three annotators for that language, giving
approximately 1{,}500 item-level annotations.

\paragraph{Validation instructions.} Annotators assessed five aspects
per item: conversation coherence and realism, epistemic-state clarity,
strategy-label correctness, foil-type accuracy, and response
acceptability. Items were presented with progressive disclosure: the
persona summary, friction traits, prior-session summaries and recent
context were shown first, and the four candidate responses were revealed
only after the annotator had committed to the state judgments, so that
the responses could not leak the state label. The four responses were
shuffled per item. The rubric was:

\begin{enumerate}\itemsep2pt
\item \textbf{Conversation coherence and realism.} ``Reading only the
user's side, does this read like a real person continuing a real
relationship with an assistant --- consistent with who they were in
earlier sessions, and with plausible reasons for what they raise now?''
Rated on a 3-point scale (fully coherent / minor lapse / breaks
character), with a free-text field required for anything below the top
rating.
\item \textbf{Epistemic-state clarity.} ``Based only on what the user
has said across these sessions, how well does the assistant know the
preference at issue in this turn?'' Options: the user has clearly stated
it and not changed it (established); the user has hinted at it or been
ambiguous (uncertain); the user has previously stated one thing and has
since signalled something different (revised); cannot tell. Annotators
were also asked to mark whether the current turn itself signals a change.
\item \textbf{Strategy-label correctness.} For each of the four
responses: ``What conversational strategy does this response use?''
Annotators selected one of the five strategies directly ---
\emph{exploit} (acts on what it already knows), \emph{elicit} (asks
an open-ended question to gather missing information),
\emph{confirm} (verifies a tentative preference before acting),
\emph{hedge} (offers options while flagging uncertainty), or
\emph{recalibrate} (acknowledges a changed preference and updates
accordingly) --- each accompanied by a one-line definition in the annotator codebook.
\item \textbf{Foil-type accuracy.} For each response judged
unacceptable: ``Which problem applies \emph{first}?'', choosing from the
seven foil types in a fixed priority order, or ``no problem''. Forcing a
single first-problem choice, rather than multi-labelling, is what makes
this measurable: several foils exhibit secondary flaws, and free
multi-selection collapses agreement.
\item \textbf{Response acceptability.} Acceptability is derived rather
than asked directly: a response is acceptable iff the annotator marked
``no problem'' at step 4. Annotators additionally selected the single
best of the four responses, which yields the human MCQ accuracy reported
below.
\end{enumerate}

\begin{table}[ht]
\centering
\small
\caption{Inter-annotator agreement by validation attribute (Cohen's
$\kappa$ between majority human label and LLM jury label, pooled
across all five languages).}
\label{tab:app_agreement_by_attribute}
\begin{tabular}{lc}
\toprule
\textbf{Attribute} & \textbf{$\kappa$} \\
\midrule
Conversation coherence and realism   & 0.84 \\
Epistemic-state clarity              & 0.76 \\
Preference-change signal (per turn)  & 0.85 \\
Strategy-label correctness (5-way)   & 0.71 \\
Foil-type accuracy (7-way)           & 0.69 \\
\midrule
Response acceptability   & 0.78 \\
\bottomrule
\end{tabular}
\end{table}

Table~\ref{tab:app_agreement_by_attribute} reports agreement separately
for each of the five judgments above; agreement is highest for
near-binary, low-inference judgments (coherence, preference-change
signal) and lowest for the more granular multi-way judgments (strategy
label, foil type), consistent with the acceptability-derived $\kappa$
of 0.78 sitting between these extremes.

\begin{table}[ht]
\centering
\small
\caption{Human validation results by language. $\kappa$ is Cohen's
$\kappa$ between the majority human label and the LLM jury label on the
acceptability judgment; human MCQ accuracy is the rate at which the
annotator's best-of-four pick was an acceptable response.}
\label{tab:app_human_validation_lang}
\begin{tabular}{lccc}
\toprule
\textbf{Language} & \textbf{$\kappa$ (vs.\ LLM jury)} & \textbf{Human MCQ acc.} & \textbf{\# items} \\
\midrule
English    & 0.81 & 96\% & 100 \\
Spanish    & 0.80 & 94\% & 100 \\
Portuguese & 0.78 & 94\% & 100 \\
Hindi      & 0.75 & 89\% & 100 \\
Indonesian & 0.76 & 87\% & 100 \\
\midrule
Overall    & 0.78 & 92\% & 500 \\
\bottomrule
\end{tabular}
\end{table}

Agreement is substantial and stable across languages, with the two
lower-resource languages (Hindi, Indonesian) trailing English by
0.05--0.06 $\kappa$. Human MCQ accuracy of 92\% sits above most
evaluated models, confirming that the benchmark is vastly
solvable and that the gap we report is a model limitation rather than
label noise.

\paragraph{Acceptable-strategy partition search.} Table~\ref{tab:app_partition_search}
reports the full ranked results of the two-group partition search
described in \S\ref{sec:human_study}, for each epistemic state. For each
of the $2^{5-1}-1 = 15$ distinct two-group partitions of the five
strategies, we fit a two-rate model in which every strategy within a
group shares a common preference rate, and score the partition by the
residual $\chi^2$ lack-of-fit,
$\sum_i (o_i - m_{g(i)})^2 / m_{g(i)}$, where $m_g$ is the mean observed
count in group $g$. Lower is better: the best partition is the one for
which ``preferred'' and ``dispreferred'' are internally homogeneous. In
every state the best-fitting partition recovers exactly the acceptable
set marked in Table~\ref{tab:human_pref}.

\begin{table}[ht]
\centering
\small
\caption{Top candidate partitions by $\chi^2$ fit, per epistemic
state. Rank 1 is the best-fitting partition. The listed set is the
\emph{preferred} group; its complement is the dispreferred group.}
\label{tab:app_partition_search}
\begin{tabular}{clccc}
\toprule
\textbf{State} & \textbf{Rank} & \textbf{Partition} & $\chi^2$ & \textbf{Matches Table~\ref{tab:human_pref}?} \\
\midrule
Established & 1 & \{Exploit\} & 1.8 & Yes \\
            & 2 & \{Exploit, Hedge\} & 419.3 & No \\
            & 3 & \{Exploit, Elicit\} & 429.0 & No \\
\midrule
Uncertain   & 1 & \{Elicit, Confirm, Hedge\} & 1.0 & Yes \\
            & 2 & \{Elicit, Confirm, Hedge, Recalibrate\} & 48.1 & No \\
            & 3 & \{Exploit, Elicit, Confirm, Hedge\} & 57.8 & No \\
\midrule
Revised     & 1 & \{Elicit, Confirm, Hedge, Recalibrate\} & 4.0 & Yes \\
            & 2 & \{Confirm, Hedge, Recalibrate\} & 13.0 & No \\
            & 3 & \{Confirm, Recalibrate\} & 20.8 & No \\
\bottomrule
\end{tabular}
\end{table}

\subsection{Comparison to Prior Personalization Resources}
\label{app:dataset_comparison}

Table~\ref{tab:app_comparison} compares our benchmark to prior
personalization resources along the axes most relevant to our claims.
\emph{Oracle state} means the resource ships a ground-truth label for
what the assistant is entitled to believe at each point, rather than
only what the user prefers. \emph{Recognition / generation split} means
the same items support both a forced-choice and a free-response
evaluation, so that recognition and generation failures can be
separated. Only HorizonBench provides a comparable oracle label; unlike our
epistemic-state framework, it labels \emph{when} a preference value
changed rather than whether the assistant is currently entitled to
act on it, hedge, or seek confirmation---the distinction our
acceptable-strategy sets operationalize.

\begin{table}[ht]
\centering
\small
\caption{Comparison to prior personalization benchmarks.}
\label{tab:app_comparison}
\begin{tabular}{lccccc}
\toprule
\textbf{Resource} & \textbf{Oracle} & \textbf{Multi-} & \textbf{Recognition /} & \textbf{Human} & \textbf{Lang.} \\
 & \textbf{state} & \textbf{session} & \textbf{generation split} & \textbf{validated} & \\
\midrule
LaMP         & \texttimes & \texttimes & \texttimes & \texttimes & 1 \\
PersonalLLM  & \texttimes & \texttimes & \texttimes & \texttimes & 1 \\
LongMemEval  & \texttimes & \checkmark & \texttimes & \checkmark & 1 \\
HorizonBench & \checkmark & \checkmark & \texttimes & \checkmark & 1 \\
PersonaLens  & \texttimes & \checkmark & \texttimes & \checkmark & 1 \\
AlpsBench  \cite{xiao2026alpsbench}  & \texttimes & \checkmark & \texttimes & \texttimes & 1 \\
\midrule
\textbf{Ours} & \checkmark & \checkmark & \checkmark & \checkmark & 5 \\
\bottomrule
\end{tabular}
\end{table}

\section{Results}
Here, we detail additional LLM results for our task. 

\subsection{Full Statistical Specification}
\label{app:statistics}

\paragraph{Model specification.} Strategy calibration (SC) by
epistemic state is modeled via mixed-effects logistic regression,
with turns nested within conversations nested within personas:
\[
\text{SC}_{ijk} \sim \text{Bernoulli}(p_{ijk}), \quad
\text{logit}(p_{ijk}) = \beta_0 + \beta_1 \text{State}_{ijk} + u_j + v_{j(k)}
\]
where $u_j \sim \mathcal{N}(0, \sigma_u^2)$ is a random intercept for
persona $j$, and $v_{j(k)} \sim \mathcal{N}(0, \sigma_v^2)$ a random
intercept for conversation $k$ nested within persona $j$. State is
dummy-coded with \textsc{Established} as the reference level.

\begin{table}[ht]
\centering
\small
\caption{\textbf{Fixed-effect estimates for the state effect on
strategy calibration.}}
\label{tab:app_mixed_effects}
\begin{tabular}{lcccc}
\toprule
\textbf{Term} & $\boldsymbol{\beta}$ & \textbf{SE} & \textbf{95\% CI} & $\boldsymbol{p}$ \\
\midrule
Intercept (Established) & 1.59 & 0.07 & [1.45, 1.73] & $<.001$ \\
Uncertain & $-2.16$ & 0.09 & [$-2.34$, $-1.98$] & $<.001$ \\
Revised & $-1.67$ & 0.09 & [$-1.85$, $-1.49$] & $<.001$ \\
\bottomrule
\end{tabular}
\end{table}

\begin{table}[ht]
\centering
\footnotesize
\caption{\textbf{Free-response performance by language and epistemic state}, averaged across the eight evaluated models. SC = strategy calibration (\%), R = rubric quality (out of 5), UA = user acceptability (\%).}
\label{tab:app_per_language}
\setlength{\tabcolsep}{4pt}
\begin{tabular}{l|ccc|ccc|ccc}
\toprule
& \multicolumn{3}{c|}{\textbf{SC (\%)}} 
& \multicolumn{3}{c|}{\textbf{R}} 
& \multicolumn{3}{c}{\textbf{UA (\%)}} \\
\textbf{Language}
& \textbf{Est.} & \textbf{Unc.} & \textbf{Rev.}
& \textbf{Est.} & \textbf{Unc.} & \textbf{Rev.}
& \textbf{Est.} & \textbf{Unc.} & \textbf{Rev.} \\
\midrule
English    & 84.0 & 39.7 & 50.7 & 3.06 & 2.43 & 2.58 & 60.0 & 46.4 & 50.3 \\
Spanish    & 83.6 & 38.3 & 49.5 & 3.04 & 2.39 & 2.54 & 59.6 & 45.4 & 49.3 \\
Portuguese & 83.5 & 39.0 & 50.0 & 3.03 & 2.41 & 2.56 & 59.5 & 45.9 & 49.7 \\
Hindi      & 82.1 & 33.1 & 43.8 & 2.92 & 2.23 & 2.38 & 57.9 & 41.7 & 46.2 \\
Indonesian & 82.3 & 34.3 & 45.5 & 2.95 & 2.29 & 2.44 & 58.0 & 42.6 & 47.0 \\
\midrule
\textbf{All}
& \textbf{83.1} & \textbf{36.9} & \textbf{47.9}
& \textbf{3.00} & \textbf{2.35} & \textbf{2.50}
& \textbf{59.0} & \textbf{44.4} & \textbf{48.5} \\
\bottomrule
\end{tabular}
\end{table}

\paragraph{Per-language estimates.} A $\text{state}\times\text{language}$
interaction term does not significantly improve model fit
(likelihood-ratio test, $\chi^2(8)=11.3$, $p=.185$), indicating the
shape of the collapse is consistent across languages. Table~\ref{tab:app_per_language_fit}
reports the state effect fit separately within each language.

\begin{table}[ht]
\centering
\small
\caption{\textbf{Per-language state effect estimates.} Fit
separately within each language; all coefficients significant at
$p<.001$.}
\label{tab:app_per_language_fit}
\begin{tabular}{lcccc}
\toprule
\textbf{Language} & $\boldsymbol{\beta}$ \textbf{(Uncertain)} & \textbf{SE} & $\boldsymbol{\beta}$ \textbf{(Revised)} & \textbf{SE} \\
\midrule
English     & $-2.05$ & 0.19 & $-1.58$ & 0.20 \\
Spanish     & $-2.11$ & 0.20 & $-1.63$ & 0.19 \\
Portuguese  & $-2.08$ & 0.21 & $-1.61$ & 0.21 \\
Hindi       & $-2.34$ & 0.22 & $-1.86$ & 0.23 \\
Indonesian  & $-2.29$ & 0.21 & $-1.79$ & 0.22 \\
\bottomrule
\end{tabular}
\end{table}

\paragraph{Per-model breakdown.} Table~\ref{tab:app_per_model}
reports free-response strategy calibration by state for each of the
eight evaluated models, underlying the closed-vs.-open-weight margin
reported in \S\ref{sec:collapse} (4.2pp Established, 16.9pp
Uncertain, 25.3pp Revised).

\begin{table}[ht]
\centering
\small
\caption{\textbf{Free-response strategy calibration (\%) by model and epistemic state.}}
\label{tab:app_per_model}
\begin{tabular}{lccc}
\toprule
\textbf{Model} & \textbf{Established} & \textbf{Uncertain} & \textbf{Revised} \\
\midrule
GPT-5.4       & 90 & 62 & 68 \\
Opus-4.7      & 77 & 45 & 66 \\
Grok-4.1      & 89 & 39 & 56 \\
Gemini-3-Pro  & 85 & 35 & 52 \\
\midrule
Gemma-3-12B   & 85 & 29 & 38 \\
Llama-3.1-8B  & 84 & 34 & 48 \\
OLMo-3-7B     & 80 & 28 & 31 \\
Qwen3-8B      & 75 & 22 & 24 \\
\midrule
Closed-weight mean & 85.2 & 45.2 & 60.5 \\
Open-weight mean   & 81.0 & 28.2 & 35.2 \\
\bottomrule
\end{tabular}
\end{table}
\subsection{Recognition--Generation Gap: Full Specification}
\label{app:recognition_generation}

\paragraph{Model specification.} The method$\times$state interaction
is fit via mixed-effects logistic regression predicting correctness
from method (MCQ vs.\ FRG), state, and their interaction, with random
intercepts for persona and conversation nested within persona
(reference levels: MCQ, Established).

\begin{table}[ht]
\centering
\small
\caption{\textbf{Fixed-effect estimates for the method$\times$state
interaction.}}
\label{tab:app_method_state}
\begin{tabular}{lcccc}
\toprule
\textbf{Term} & $\boldsymbol{\beta}$ & \textbf{SE} & \textbf{95\% CI} & $\boldsymbol{p}$ \\
\midrule
FRG (main effect) & $-0.31$ & 0.08 & [$-0.47$, $-0.15$] & $<.001$ \\
Uncertain (main effect) & $-0.44$ & 0.07 & [$-0.58$, $-0.30$] & $<.001$ \\
Revised (main effect) & $-0.38$ & 0.07 & [$-0.52$, $-0.24$] & $<.001$ \\
FRG $\times$ Uncertain & $-1.82$ & 0.10 & [$-2.02$, $-1.62$] & $<.001$ \\
FRG $\times$ Revised & $-1.68$ & 0.10 & [$-1.88$, $-1.48$] & $<.001$ \\
\bottomrule
\end{tabular}
\end{table}

\begin{table}[ht]
\centering
\small
\caption{\textbf{MCQ strategy calibration (\%) by model and
epistemic state.}}
\label{tab:app_mcq_per_model}
\begin{tabular}{lccc}
\toprule
\textbf{Model} & \textbf{Established} & \textbf{Uncertain} & \textbf{Revised} \\
\midrule
GPT-5.4       & 95 & 90 & 91 \\
Opus-4.7      & 91 & 87 & 89 \\
Grok-4.1      & 94 & 86 & 87 \\
Gemini-3-Pro  & 92 & 84 & 85 \\
Gemma-3-12B   & 90 & 81 & 82 \\
Llama-3.1-8B  & 89 & 79 & 80 \\
OLMo-3-7B     & 87 & 77 & 78 \\
Qwen3-8B      & 84 & 75 & 76 \\
\midrule
Mean & 90.3 & 82.4 & 83.5 \\
\bottomrule
\end{tabular}
\end{table}

\subsection{Oracle-Strategy FRG: Full Results}
\label{app:oracle_strategy}

In the oracle-strategy condition (Table \ref{tab:app_oracle_strategy}), the model is given the acceptable
strategy explicitly (e.g., ``respond using \textsc{hedge}'') and
asked to generate a response accordingly, isolating execution from
selection. Models comply with the supplied strategy on 95\% of turns
overall (range 91-94\% across the four strategies tested beyond
\textsc{exploit}).

\begin{table}[ht]
\centering
\small
\caption{\textbf{Unaided vs.\ oracle-strategy performance by epistemic state.} Unaided values match \S3.1 and Table~\ref{tab:app_sc_r_ua_full}.}
\label{tab:app_oracle_strategy}
\begin{tabular}{lccc}
\toprule
\textbf{Metric} & \textbf{Established} & \textbf{Uncertain} & \textbf{Revised} \\
\midrule
SC, unaided (\%)          & 83.1 & 36.9 & 47.9 \\
SC, oracle-strategy (\%)  & 96.4 & 93.1 & 94.0 \\
Rubric, unaided           &  3.00 &  2.35 &  2.50 \\
Rubric, oracle-strategy   & 3.58 & 3.52 & 3.56 \\
UA, unaided (\%)          &  59.0 &  44.4 &  48.5 \\
UA, oracle-strategy (\%)  & 60.2 & 58.7 & 59.3 \\
\bottomrule
\end{tabular}
\end{table}

\paragraph{Execution quality by strategy, not just by state.} To
confirm the state-invariance of oracle-strategy execution is not an
artifact of some strategies being intrinsically easier than others,
Table~\ref{tab:app_oracle_by_strategy} reports rubric quality broken
down by the supplied strategy.

\begin{table}[ht]
\centering
\small
\caption{\textbf{Oracle-strategy execution quality (Rubric, Acc/5)
by supplied strategy.} Compliance rate is the fraction of turns on
which the model produced a response matching the supplied strategy.}
\label{tab:app_oracle_by_strategy}
\begin{tabular}{lcc}
\toprule
\textbf{Strategy} & \textbf{Compliance (\%)} & \textbf{Rubric (Acc/5)} \\
\midrule
Exploit     & 98 & 3.61 \\
Elicit      & 93 & 3.49 \\
Confirm     & 94 & 3.55 \\
Hedge       & 91 & 3.44 \\
Recalibrate & 92 & 3.47 \\
\bottomrule
\end{tabular}
\end{table}

\subsection{Generation Strategy Distribution: Full Breakdown}

\begin{figure*}[t]
\centering
\includegraphics[width=\textwidth]{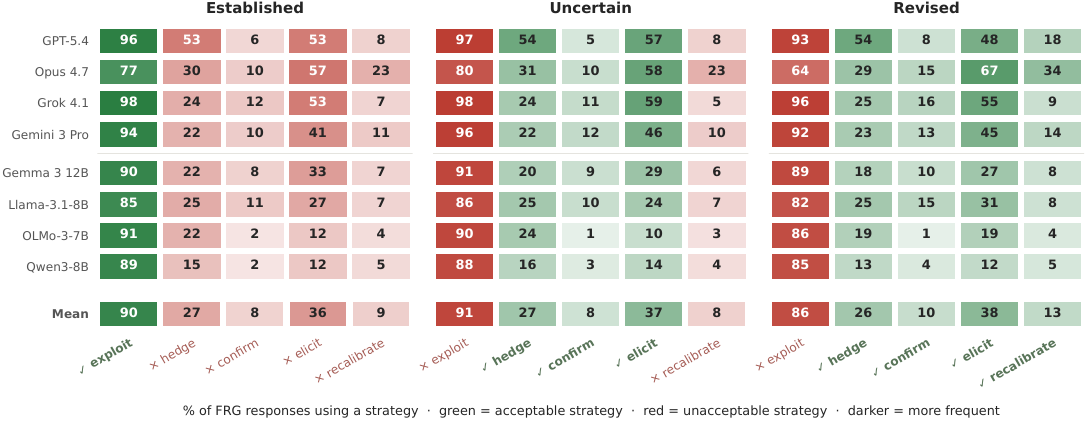}
\caption{\textbf{Strategy presence in free-response generations.} Percentage
of responses containing each strategy (multi-label: a response can use
several strategies, so rows need not sum to 100\%), pooled across five
languages. Exploitation appears in nearly all responses across states,
frequently alongside \textit{elicit} or \textit{hedge}. The strict and lenient
scores in Table~\ref{tab:app_scoring_rules} are computed from these tags.}
\label{fig:app_presence}
\end{figure*}

\paragraph{Foil-type contingency table.} Table~\ref{tab:app_foil_contingency}
reports the full distribution underlying the reported $\chi^2$ test
of independence ($\chi^2(12)=1332.7$, $p<.001$).

\begin{table}[ht]
\centering
\small
\caption{\textbf{Foil-type distribution (\%) by epistemic state.}}
\label{tab:app_foil_contingency}
\begin{tabular}{lccc}
\toprule
\textbf{Foil type} & \textbf{Established} & \textbf{Uncertain} & \textbf{Revised} \\
\midrule
Over-personalization & 31.2 & 9.2  & 7.3  \\
Epistemic overreach  & 20.7 & 16.7 & 14.6 \\
Staleness            & 0.0  & 10.7 & 29.9 \\
Need mismatch        & 9.8  & 8.2  & 9.8  \\
Over-probing         & 21.4 & 6.9  & 6.9  \\
Style mismatch       & 8.7  & 5.6  & 5.8  \\
Failure to probe     & 8.2  & 42.7 & 25.7 \\
\bottomrule
\end{tabular}
\end{table}

\paragraph{Per-language foil-type breakdown.} Table~\ref{tab:app_per_lang_foil}
reports the dominant foil type under \textsc{Uncertain} (the state
showing the largest cross-language variation) for each language.

\begin{table}[ht]
\centering
\small
\caption{\textbf{Dominant foil type under \textsc{Uncertain}, by
language.}}
\label{tab:app_per_lang_foil}
\begin{tabular}{lcc}
\toprule
\textbf{Language} & \textbf{Failure-to-probe (\%)} & \textbf{Error rate (\%)} \\
\midrule
English     & 36.2 & 60.3 \\
Spanish     & 37.1 & 61.7 \\
Portuguese  & 35.8 & 61.0 \\
Hindi       & 43.5 & 66.9 \\
Indonesian  & 42.1 & 65.7 \\
\bottomrule
\end{tabular}
\end{table}

\subsection{Sensitivity to Strategy-Calibration Scoring Rule}
\label{app:scoring_rules}

Because model responses often combine strategies (Figure~\ref{fig:app_presence}),
we recompute SC under two alternative rules: a \textit{strict} rule (only
acceptable strategies present) and a \textit{lenient} rule (any acceptable
strategy present). The collapse persists under lenient scoring: mean SC falls
from 90.0\% (\textsc{Established}) to 49.1\% (\textsc{Uncertain}) and 57.4\%
(\textsc{Revised}), gaps modestly narrower than under dominant scoring
(40.9 vs.\ 46.2 pp and 32.6 vs.\ 35.2 pp). Strict SC is near zero outside
\textsc{Established}, as nearly all responses contain some exploitation, and
only 39.5\% under \textsc{Established}, reflecting that models append questions
or hedges at a similar rate in every state (Figure~\ref{fig:app_presence}),
consistent with a state-invariant policy (\S3.3). Under lenient scoring, the
recognition--generation gap under \textsc{Established} largely disappears,
while the gaps under \textsc{Uncertain} and \textsc{Revised} remain large
(33 and 26 pp). Lenient scoring credits strategy stacking regardless of state,
so we treat it as an upper bound rather than a primary metric
(Table~\ref{tab:app_scoring_rules}).

\begin{table}[ht]
\centering
\small
\caption{\textbf{Strategy calibration (\%) under alternative scoring rules.}
\textit{Strict}: only acceptable strategies present; \textit{Dom.}: dominant
strategy acceptable (main metric); \textit{Len.}: any acceptable strategy
present. By construction, Strict $\leq$ Dom.\ $\leq$ Len.}
\label{tab:app_scoring_rules}
\setlength{\tabcolsep}{4pt}
\begin{tabular}{l ccc ccc ccc}
\toprule
& \multicolumn{3}{c}{\textbf{Established}}
& \multicolumn{3}{c}{\textbf{Uncertain}}
& \multicolumn{3}{c}{\textbf{Revised}} \\
\cmidrule(lr){2-4} \cmidrule(lr){5-7} \cmidrule(lr){8-10}
\textbf{Model} & Strict & Dom. & Len. & Strict & Dom. & Len. & Strict & Dom. & Len. \\
\midrule
GPT-5.4      & 41 & 90 & 96 & 5 & 62 & 66 & 6 & 68 & 72 \\
Opus-4.7     & 32 & 77 & 77 & 4 & 45 & 58 & 7 & 66 & 73 \\
Grok-4.1     & 42 & 89 & 98 & 3 & 39 & 64 & 5 & 56 & 68 \\
Gemini-3-Pro & 38 & 85 & 94 & 4 & 35 & 58 & 6 & 52 & 65 \\
\midrule
Gemma-3-12B  & 35 & 85 & 90 & 3 & 29 & 39 & 5 & 38 & 45 \\
Llama-3.1-8B & 37 & 84 & 85 & 3 & 34 & 40 & 6 & 48 & 51 \\
OLMo-3-7B    & 40 & 80 & 91 & 2 & 28 & 31 & 4 & 31 & 37 \\
Qwen3-8B     & 39 & 75 & 89 & 2 & 22 & 29 & 3 & 24 & 31 \\
\midrule
Mean         & 38.0 & 83.1 & 91.0
             & 3.3 & 36.9 & 48.1
             & 5.3 & 47.9 & 55.3 \\
\bottomrule
\end{tabular}
\end{table}

\subsection{Strategy Calibration and Response Quality: Full Data}
\label{app:sc_r_ua_full}

\paragraph{Per-model, per-state data underlying within-state
correlations.} Table~\ref{tab:app_sc_r_ua_full} reports the raw
SC, R, and UA values for each model and epistemic state used to
compute the correlations reported in \S\ref{sec:sc_r_ua}.

\begin{table}[ht]
\centering
\small
\caption{\textbf{SC, R, and UA by model and epistemic state.}}
\label{tab:app_sc_r_ua_full}
\begin{tabular}{llccc}
\toprule
\textbf{State} & \textbf{Model} & \textbf{SC (\%)} & \textbf{R} & \textbf{UA (\%)} \\
\midrule
\multirow{8}{*}{Established}
& GPT-5.4      & 90 &   {4.29} &   {95} \\
& Opus-4.7     & 77 &   {2.88} &   {72} \\
& Grok-4.1     & 89 &   {2.67} &   {71} \\
& Gemini-3-Pro & 85 &   {3.16} &   {63} \\
& Gemma-3-12B  & 85 &   {3.38} &   {52} \\
& Llama-3.1-8B & 84 &   {3.00} &   {51} \\
& OLMo-3-7B    & 80 &   {2.53} &   {43} \\
& Qwen3-8B     & 75 &   {2.10} &   {25} \\
\midrule
\multirow{8}{*}{Uncertain}
& GPT-5.4      & 62 &   {3.38} &   {52} \\
& Opus-4.7     & 45 &   {2.01} &   {66} \\
& Grok-4.1     & 39 &   {1.85} &   {58} \\
& Gemini-3-Pro & 35 &   {2.59} &   {57} \\
& Gemma-3-12B  & 29 &   {3.04} &   {47} \\
& Llama-3.1-8B & 34 &   {2.42} &   {26} \\
& OLMo-3-7B    & 28 &   {1.97} &   {34} \\
& Qwen3-8B     & 22 &   {1.57} &   {15} \\
\midrule
\multirow{8}{*}{Revised}
& GPT-5.4      & 68 &   {3.79} &   {93} \\
& Opus-4.7     & 66 &   {2.43} &   {57} \\
& Grok-4.1     & 56 &   {2.17} &   {51} \\
& Gemini-3-Pro & 52 &   {2.66} &   {45} \\
& Gemma-3-12B  & 38 &   {2.86} &   {45} \\
& Llama-3.1-8B & 48 &   {2.50} &   {28} \\
& OLMo-3-7B    & 31 &   {2.02} &   {43} \\
& Qwen3-8B     & 24 &   {1.59} &   {26} \\
\bottomrule
\end{tabular}
\end{table}

\paragraph{Note on statistical power.} Each within-state correlation
is computed across eight models ($n=8$); at this sample size,
correlation estimates carry wide confidence intervals and should be
interpreted as descriptive rather than as precisely-powered
inferential estimates. We report them because the direction and
relative ordering across states (strongest under Established and
Revised, weaker under Uncertain) is consistent and theoretically
motivated, not because individual coefficients should be treated as
tightly estimated.

\subsection{Intervention Prompt Templates}
\label{app:interventions}

For each intervention, the model receives the persona $P$,
conversation history $C$, and current user turn $u$ exactly as in
the unaided condition (\S\ref{sec:eval}), with the following
addition prepended or appended as noted.

\paragraph{(1) Explicit strategy instructions.} Prepended to the
system prompt:

\begin{quote}
\small\ttfamily
When responding, choose one of the following five conversational
strategies based on how reliable the user's stated preference
currently is:\\
- \textbf{Exploit}: act directly on a preference that is clearly
established and uncontradicted.\\
- \textbf{Elicit}: ask an open-ended question when you have no
information about the relevant preference.\\
- \textbf{Confirm}: verify a preference that has been stated but
remains tentative or ambiguous.\\
- \textbf{Hedge}: offer options without fully committing, when
evidence is mixed or uncertain.\\
- \textbf{Recalibrate}: explicitly acknowledge and update when new
evidence contradicts a previously stated preference.\\
Select the strategy that best matches the current state of the
user's preference, then respond accordingly.
\end{quote}

\paragraph{(2) Oracle preference memory.} Prepended before the
conversation history, a structured summary generated from the
benchmark's oracle labels:

\begin{quote}
\small\ttfamily
Here is a summary of what you currently know about this user's
preferences:\\
- \{preference\_key\}: \textbf{established} --- the user has stated
this clearly and it has not been contradicted.\\
- \{preference\_key\}: \textbf{uncertain} --- you have limited or
ambiguous evidence about this.\\
- \{preference\_key\}: \textbf{revised} --- the user has indicated
this preference has changed from what was previously stated.\\
Use this summary to guide how confidently you act on each
preference.
\end{quote}

\paragraph{(3) Chain-of-thought (CoT).} Appended after the user
turn, following \citet{wei2022chain}:

\begin{quote}
\small\ttfamily
Before responding, think step by step: (1) What preference is
relevant to this turn? (2) How reliable is the evidence for that
preference right now---established, uncertain, or revised? (3)
Given that, what conversational strategy is appropriate? Then
provide your response.
\end{quote}

\paragraph{(4) Few-shot examples.} Three worked examples prepended
before the conversation history, one per epistemic state, in the
same format:

\begin{quote}
\small\ttfamily
Example (established preference): \\
User: [earlier turn stating a clear preference] \\
User (current turn): [unrelated follow-up] \\
Assistant (exploit): [confident response acting on the preference] \\[4pt]
Example (uncertain preference): \\
User: [ambiguous or partial signal about a preference] \\
Assistant (elicit/confirm/hedge): [response managing the uncertainty] \\[4pt]
Example (revised preference): \\
User: [statement contradicting an earlier stated preference] \\
Assistant (recalibrate): [response explicitly updating]
\end{quote}

\paragraph{(5) Best-of-five sampling.} No prompt modification; five
independent generations are sampled using temperature=0.7 for the same $(P, C, u)$, and the response maximizing an equally-weighted sum of
SC, R, and UA (as scored by the LLM jury / user simulator,
\S\ref{sec:benchmark_construct}) is retained.


\begin{table}[ht]
\centering
\small
\caption{\textbf{Closed-weight vs.\ open-weight intervention effects}, averaged across all five interventions and all three epistemic states.}
\label{tab:app_openweight_intervention}
\begin{tabular}{lccc}
\toprule
\textbf{Model class} & $\boldsymbol{\Delta}$\textbf{SC (pp)} & $\boldsymbol{\Delta}$\textbf{R} & $\boldsymbol{\Delta}$\textbf{UA (pp)} \\
\midrule
Closed-weight &  {$+1.4$} &  {$+0.04$} &  {$+0.46$} \\
Open-weight   &  {$+0.9$} &  {$+0.03$} &  {$+0.32$} \\
\bottomrule
\end{tabular}
\end{table}

\begin{table}[ht]
\centering
\small
\caption{\textbf{Intervention effects by language}, averaged across all five interventions and all three epistemic states.}
\label{tab:app_language_intervention}
\begin{tabular}{lccc}
\toprule
\textbf{Language} & $\boldsymbol{\Delta}$\textbf{SC (pp)} & $\boldsymbol{\Delta}$\textbf{R} & $\boldsymbol{\Delta}$\textbf{UA (pp)} \\
\midrule
English    &  {$+1.3$} &  {$+0.05$} &  {$+0.46$} \\
Spanish    &  {$+1.3$} &  {$+0.04$} &  {$+0.44$} \\
Portuguese &  {$+1.2$} &  {$+0.04$} &  {$+0.43$} \\
Hindi      &  {$+1.0$} &  {$+0.03$} &  {$+0.30$} \\
Indonesian &  {$+1.0$} &  {$+0.03$} &  {$+0.32$} \\
\bottomrule
\end{tabular}
\end{table}

\begin{table}[ht]
\centering
\small
\caption{\textbf{Change in SC, R, and UA relative to baseline, by intervention and epistemic state.} Averaged across eight models and five languages.}
\label{tab:app_intervention_full}
\begin{tabular}{llccc}
\toprule
\textbf{Intervention} & \textbf{State} & $\boldsymbol{\Delta}$\textbf{SC (pp)} & $\boldsymbol{\Delta}$\textbf{R} & $\boldsymbol{\Delta}$\textbf{UA (pp)} \\
\midrule
\multirow{3}{*}{Explicit instructions}
& Established & $-0.8$ & $+0.01$ & $-0.65$ \\
& Uncertain   & $+3.0$ & $+0.05$ & $+1.14$ \\
& Revised     & $+3.5$ & $+0.15$ & $+1.47$ \\
\midrule
\multirow{3}{*}{Oracle memory}
& Established & $-0.6$ & $-0.02$ & $-0.43$ \\
& Uncertain   & $+0.4$ & $+0.01$ & $+0.02$ \\
& Revised     & $+0.8$ & $-0.02$ & $+0.59$ \\
\midrule
\multirow{3}{*}{CoT}
& Established & $-1.5$ & $-0.02$ & $-0.09$ \\
& Uncertain   & $+2.7$ & $+0.08$ & $+0.96$ \\
& Revised     & $+3.4$ & $+0.09$ & $+0.90$ \\
\midrule
\multirow{3}{*}{Few-shot}
& Established & $-3.4$ & $-0.05$ & $-1.23$ \\
& Uncertain   & $+3.5$ & $+0.11$ & $+0.84$ \\
& Revised     & $+4.7$ & $+0.13$ & $+1.77$ \\
\midrule
\multirow{3}{*}{Best-of-five}
& Established & $+0.2$ & $+0.01$ & $+0.10$ \\
& Uncertain   & $+0.6$ & $+0.02$ & $+0.20$ \\
& Revised     & $+0.8$ & $+0.02$ & $+0.25$ \\
\bottomrule
\end{tabular}
\end{table}

\subsection{Training Details}
\label{app:training}

\paragraph{Training and held-out data construction.} We generate a
training corpus using the same three-role pipeline as our benchmark
(\S\ref{sec:benchmark_construct}) using open-weight models (Kimi-K2.6, Qwen3-235B, and DeepSeek-V3), but from a disjoint set of 144
personas per language (720 total across five languages), entirely
non-overlapping with the 250 personas used in our benchmark
(\S\ref{sec:task}). This yields approximately 5k conversational
turns per language (24k total), spanning the same controlled
epistemic-state transitions (established, uncertain, revised). We
hold out 10\% of this training corpus (persona-disjoint from the
training split) as an \emph{in-distribution} evaluation set, used to
compute training-time performance and the in-distribution half of the
OOD generalization gap reported in \S\ref{sec:training}. The
25,006-turn benchmark itself (\S\ref{sec:task}) is never used during
training and serves as the out-of-distribution evaluation set for
all reported generalization results.

\paragraph{Base model fine-tuning (SFT, DPO).} All fine-tuning uses
LoRA \citep{hu2021lora} with rank $r=16$, $\alpha=32$, dropout 0.05,
applied to query, key, value, and output projection matrices. SFT
trains for 3 epochs with AdamW, learning rate $2\times10^{-5}$,
cosine schedule with 3\% warmup, effective batch size 32 (per-device
batch 4, gradient accumulation 8) with acceptable responses from the training corpus as the prediction target. DPO initializes from the SFT
checkpoint, trains for 2 epochs, learning rate $5\times10^{-6}$,
$\beta=0.1$, using acceptable-vs-foil response pairs from the
training corpus as preference pairs.

\paragraph{GRPO training (base, \textsc{Link}, \textsc{Link+CF}).}
All GRPO configurations initialize from the SFT checkpoint and use
group size $G=8$ rollouts per prompt, learning rate
$1\times10^{-6}$, KL penalty coefficient $0.02$ against the SFT
reference policy, and PPO-style clipping with $\epsilon=0.2$. Each
configuration trains for 800 policy update steps, with 64 prompts per
batch (512 total rollouts per step). The base reward
$r_{\text{base}} = \tfrac{1}{3}(\text{SC} + \text{UA} + \text{R})$ is
computed via the open-weight models used for training dataset generation on each rollout (disjoint from the closed-weight models used for benchmark generation/evaluation). Since R scores range from 1 to 5, we normalize these values and scale to 0 to 1 to match SC and UA. \textsc{Link} and \textsc{Link+CF} add the targeted
term at $\lambda=0.1$ (\S\ref{sec:training}), computed from the
current policy's hidden state at 70\% model depth, mean-pooled over the first eight generated tokens.

\paragraph{Probe refitting during training.} Because \textsc{Link}
and \textsc{Link+CF} depend on a strategy probe $P_\phi(\cdot\mid h)$
applied to the \emph{current} policy's hidden states, we refit both
the state and strategy probes every $N=10$ GRPO steps, using a fresh
sample of 2,000 rollouts generated by the current policy, to keep the
probe calibrated as the policy drifts during training. Probes are
logistic regression classifiers trained on frozen hidden-state
activations, following the same procedure as \S\ref{sec:training}.
This periodic refitting is necessary for the reward to remain
well-defined as training progresses.

\paragraph{Counterfactual patching for \textsc{Link+CF}.} For each
rollout, an alternate state $s' \neq s$ is sampled uniformly from the
remaining two states. We patch $h$ toward $s'$ by adding the
difference between the mean probe-direction projections of $s'$- and
$s$-labeled activations (estimated from the same 2,000-rollout sample
used for probe refitting), scaled by a coefficient $\alpha=1.0$
(full mean-difference magnitude).

\paragraph{Compute.} All experiments run on 4-8 NVIDIA A100 80GB GPUs.
Table~\ref{tab:app_compute} reports approximate compute hours per method, per model.

\begin{table}[ht]
\centering
\small
\caption{\textbf{Approximate compute per training run}, per model.
GRPO configurations are more expensive due to rollout generation.}
\label{tab:app_compute}
\begin{tabular}{lcc}
\toprule
\textbf{Method} & \textbf{Total time to train each model} \\
\midrule
SFT              & 8  \\
DPO               & 15  \\
GRPO (base)       & 45  \\ 
GRPO+\textsc{Link}    & 49  \\
GRPO+\textsc{Link+CF} & 53 \\
\midrule
\textbf{Total} & & \textbf{170} \\
\bottomrule
\end{tabular}
\end{table}

\begin{table}[ht]
\centering
\small
\caption{\textbf{Per-model results on the held-out in-distribution set}, averaged across five languages. SC = strategy calibration (\%), R = rubric (out of 5), UA = user acceptability (\%). Benchmark results are shown in Figure~\ref{fig:training}; the difference is the generalization gap reported in \S\ref{sec:training}.}
\label{tab:training_full}
\begin{tabular}{llccc}
\toprule
\textbf{Model} & \textbf{Method} & \textbf{SC} & \textbf{R} & \textbf{UA} \\
\midrule
\multirow{6}{*}{Llama-3.1-8B}
& Untrained             & 57 & 2.74 & 37 \\
& SFT                   & 70 & 3.33 & 45 \\
& DPO                   & 69 & 3.29 & 44 \\
& GRPO (base)           & 76 & 3.59 & 51 \\
& GRPO+\textsc{Link}    & 80 & 3.69 & 53 \\
& GRPO+\textsc{Link+CF} & 82 & 3.69 & 52 \\
\midrule
\multirow{6}{*}{Gemma-3-12B}
& Untrained             & 53 & 3.19 & 50 \\
& SFT                   & 65 & 3.76 & 58 \\
& DPO                   & 67 & 3.82 & 59 \\
& GRPO (base)           & 70 & 3.89 & 60 \\
& GRPO+\textsc{Link}    & 77 & 3.95 & 65 \\
& GRPO+\textsc{Link+CF} & 75 & 3.98 & 66 \\
\midrule
\multirow{6}{*}{OLMo-3-7B}
& Untrained             & 49 & 2.27 & 42 \\
& SFT                   & 60 & 2.88 & 47 \\
& DPO                   & 60 & 2.84 & 51 \\
& GRPO (base)           & 65 & 3.19 & 54 \\
& GRPO+\textsc{Link}    & 68 & 3.34 & 55 \\
& GRPO+\textsc{Link+CF} & 68 & 3.32 & 54 \\
\midrule
\multirow{6}{*}{Qwen3-8B}
& Untrained             & 43 & 1.85 & 24 \\
& SFT                   & 54 & 2.48 & 33 \\
& DPO                   & 57 & 2.68 & 32 \\
& GRPO (base)           & 57 & 2.97 & 39 \\
& GRPO+\textsc{Link}    & 63 & 3.10 & 39 \\
& GRPO+\textsc{Link+CF} & 67 & 3.21 & 44 \\
\bottomrule
\end{tabular}
\end{table}

\end{document}